\documentclass[preprint,12pt]{elsarticle}

\usepackage{url}
\usepackage{caption}
\usepackage{subcaption}
\usepackage[ruled,vlined,linesnumbered]{algorithm2e}
\usepackage{algorithmic}
\usepackage{bbding}
\usepackage[normalem]{ulem}
\useunder{\uline}{\ul}{}
\usepackage{epsfig,threeparttable,multirow,bbding,bbm, dsfont}
\usepackage{amsmath,amsfonts,amssymb,bm}
\usepackage{array}
\usepackage{textcomp}
\usepackage{stfloats}
\usepackage{placeins}
\usepackage{verbatim}
\usepackage{graphicx}
\usepackage{booktabs}
\usepackage{setspace}
\usepackage{multirow}
\usepackage{amsthm}
\usepackage{xcolor}
\usepackage{colortbl}
\usepackage{setspace}

\usepackage{hyperref}
\hypersetup{hidelinks}
\definecolor{lightkeycolor}{RGB}{255,240,240}
\definecolor{lightgray}{RGB}{234,234,234}

\newcommand{\alghl}[1]{%
\setlength{\fboxsep}{0pt}\colorbox{lightkeycolor}{#1}%
}

\newcommand{\orcidA}{\href{https://orcid.org/0000-0002-2819-0200}{}}  % Jing Liu
\newcommand{\orcidB}{\href{https://orcid.org/0000-0002-1312-0146}{}}  % Yang Liu
\newcommand{\orcidC}{\href{https://orcid.org/0000-0003-3641-1429}{}}  % Wei Zhou
\newcommand{\orcidD}{\href{https://orcid.org/0000-0002-3331-1011}{}}  % Weiping Ding
\newcommand{\orcidG}{\href{https://orcid.org/0000-0003-3529-2640}{}}  % Victor C.M. Leung
\newcommand{\orcidH}{\href{https://orcid.org/0009-0001-1913-8726}{}}  % Donglai Wei
\newcommand{\orcidI}{\href{https://orcid.org/0009-0003-8772-2187}{}}  % Sipeng Zhang
\newcommand{\orcidJ}{\href{https://orcid.org/0009-0000-7104-5394}{}}  % Tong Yang

\journal{Pattern Recognition}

\usepackage{xr}
\PassOptionsToPackage{table}{xcolor}
\def\RevisionPreambleLight{1}
\ifdefined\pdfminorversion\pdfminorversion=4\fi
\ifdefined\pdfobjcompresslevel\pdfobjcompresslevel=0\fi

\makeatletter
\PassOptionsToPackage{table}{xcolor}
\@ifpackageloaded{xcolor}{}{\usepackage{xcolor}}
\@ifpackageloaded{soul}{}{\@ifundefined{ul}{\usepackage{soul}}{}}
\@ifpackageloaded{pifont}{}{\usepackage{pifont}}
\@ifpackageloaded{tikz}{}{\usepackage{tikz}}
\makeatother
\definecolor{revisionHighlightColor}{RGB}{0,0,255}
\providecommand{\blue}[1]{}
\renewcommand{\blue}[1]{\textcolor{revisionHighlightColor}{#1}}
\providecommand{\RegisterRevisionHighlightCommands}{%
  \ifdefined\soulregister
    \ifdefined\cite\soulregister\cite7\fi
    \ifdefined\ref\soulregister\ref7\fi
    \ifdefined\cref\soulregister\cref7\fi
    \ifdefined\Cref\soulregister\Cref7\fi
    \ifdefined\eqref\soulregister\eqref7\fi
    \ifdefined\textbf\soulregister\textbf7\fi
    \ifdefined\textit\soulregister\textit7\fi
    \ifdefined\emph\soulregister\emph7\fi
  \fi
}
\RegisterRevisionHighlightCommands
\makeatletter
\@ifundefined{hlmaskenv}{%
  \ifdefined\RevisionDisableVisualHighlight
    \newenvironment{hlmaskenv}{}{}%
  \else
    \newsavebox{\hlmasksavebox}%
  \fi
}{}
\makeatother
\ifdefined\RevisionPreambleLight
  \def\RevisionPreambleMaybeEnd{ }
\else
  \let\RevisionPreambleMaybeEnd\relax
\fi
\RevisionPreambleMaybeEnd

\usepackage{mathtools,bm,amsmath,amssymb,amsfonts,amsthm,nicefrac,latexsym,dsfont,extarrows,bbm}
\usepackage{cases,subeqnarray}
\usepackage{algorithm,algorithmicx,algpseudocode}
\usepackage{booktabs,multirow,subcaption,graphicx,epsfig,epstopdf,float,stfloats,bigstrut,makecell,array}
\usepackage[most]{tcolorbox}
\usepackage{xcolor,color,colortbl,textcomp,pifont,microtype}
\usepackage{multicol,balance,flushend,setspace,titletoc}
\usepackage[noadjust]{cite}
\usepackage{url,xr-hyper}
\makeatletter
\@ifpackageloaded{wasysym}{}{\usepackage[nointegrals]{wasysym}}
\makeatother

\usepackage[pagebackref=false,breaklinks=true,colorlinks=false,pdfborder={0 0 0},bookmarks=true,linkbordercolor=white,citebordercolor=white,urlbordercolor=white,filebordercolor=white]{hyperref}
\makeatletter
\@ifundefined{Hy@setpdfversionfalse}{}{\Hy@setpdfversionfalse}
\makeatother
\usepackage[nameinlink]{cleveref}
\RegisterRevisionHighlightCommands

\usepackage{tikz}
\definecolor{lime}{HTML}{A6CE39}
\DeclareRobustCommand{\orcidicon}{\begin{tikzpicture}\draw[lime, fill=lime] (0,0) circle [radius=0.16] node[white] {{\fontfamily{qag}\selectfont \tiny ID}};\draw[white, fill=white] (-0.0625,0.095) circle [radius=0.007];\end{tikzpicture}\hspace{-2mm}}
\foreach \x in {A, ..., Z}{\expandafter\xdef\csname orcid\x\endcsname{\noexpand\href{https://orcid.org/\csname orcidauthor\x\endcsname}{\noexpand\orcidicon}}}

\theoremstyle{plain}

\theoremstyle{remark}

\crefname{section}{Sec.}{Secs.}
\crefname{subsection}{Sec.}{Secs.}
\crefname{subsubsection}{Sec.}{Secs.}
\crefname{figure}{Fig.}{Figs.}
\crefname{table}{Table}{Tables}
\crefname{equation}{Eq.}{Eqs.}
\crefname{algorithm}{Alg.}{Algs.}
\crefname{assumption}{Assumption}{Assumptions}
\crefname{theorem}{Theorem}{Theorems}
\crefname{lemma}{Lemma}{Lemmas}
\crefname{proposition}{Proposition}{Propositions}
\crefname{corollary}{Corollary}{Corollaries}
\crefname{definition}{Definition}{Definitions}
\crefformat{equation}{#2Eq.~#1#3}
\crefmultiformat{equation}{#2Eqs.~#1#3}{ and~#2#1#3}{, #2#1#3}{ and~#2#1#3}
\Crefformat{equation}{#2Eq.~#1#3}
\Crefmultiformat{equation}{#2Eqs.~#1#3}{ and~#2#1#3}{, #2#1#3}{ and~#2#1#3}

\definecolor{lightkeycolor}{RGB}{255,240,240}
\definecolor{lightgray}{RGB}{234,234,234}

\newcommand{\alghl}[1]{\setlength{\fboxsep}{0pt}\colorbox{lightkeycolor}{#1}}

\def\x{{\mathbf x}}

\ifdefined\pdfminorversion\pdfminorversion=4\fi
\ifdefined\pdfinclusioncopyfonts\pdfinclusioncopyfonts=1\fi
\IfFileExists{macros.tex}{\input{macros}}{}

\let\RevisionPreambleLight\undefined

\begin{document}

\begin{frontmatter}

\title{SCMM: Calibrating Cross-modal Representations for Text-Based Person Search}

    \author[label1,label2]{Jing Liu\orcidA{}}
    \ead{jingliu@ece.ubc.ca}
    \author[label1]{Donglai Wei\orcidH{}\corref{cor1}}
    \ead{dlwei21@m.fudan.edu.cn}
    \author[label3]{Yang Liu\orcidB{}}
    \ead{yang\_liu@ieee.org}
    \author[label4]{Sipeng Zhang\orcidI{}}
    \ead{zhangsipeng@megvii.com}
    \author[label4]{Tong Yang\orcidJ{}}
    \ead{yangtong@megvii.com}
    \author[label5]{Wei Zhou\orcidC{}}
    \ead{zhouw26@cardiff.ac.uk}
    \author[label6,label7]{Weiping Ding\orcidD{}}
    \ead{dwp9988@163.com}
    \author[label8,label2]{Victor C. M. Leung\orcidG{}}
    \ead{vleung@ece.ubc.ca}
    \cortext[cor1]{Corresponding author.}

    \affiliation[label1]{organization={College of Future Information Technology, Fudan University},
        city={Shanghai},
        postcode={200433},
        country={China}}

    \affiliation[label2]{organization={Department of ECE, The University of British Columbia},
        city={Vancouver},
        postcode={V6T 1Z4},
        country={Canada}}

    \affiliation[label4]{organization={MEGVII Technology},
        city={Beijing},
        postcode={100096},
        country={China}}

    \affiliation[label3]{organization={College of Electr. and Info. Eng., Tongji University},
        city={Shanghai},
        postcode={201804},
        country={China}}

    \affiliation[label5]{organization={School of Computer Science and Informatics, Cardiff University},
        city={Cardiff},
        postcode={CF24 4AG},
        country={U.K.}}

    \affiliation[label6]{organization={School of AI and CS, Nantong University},
        city={Nantong},
        postcode={226019},
        country={China}}

    \affiliation[label7]{organization={Faculty of Data Science, City University of Macau},
        city={Macau},
        postcode={999078},
        country={China}}

    \affiliation[label8]{organization={Academy of Artificial Intelligence, SMBU},
        city={Shenzhen},
        postcode={518115},
        state={Guangdong},
        country={China}}

\begin{abstract}
Text-Based Person Search (TBPS) aims to retrieve target person images from a large-scale database using natural language descriptions, serving as a critical task in multimodal perception and visual pattern recognition. Bridging the semantic gap between heterogeneous modalities while capturing fine-grained correspondences remains a fundamental challenge, especially when discriminating visually similar individuals based on complex textual semantics. To address these challenges, we propose Sew Calibration and Masked Modeling (SCMM), a unified framework that calibrates cross-modal representations for effective multimodal visual-textual pattern matching. Concretely, SCMM introduces two principal components: a sew calibration loss that dynamically aligns image-text features via a quality-guided adaptive margin governed by textual information density, and a masked caption modeling loss that establishes fine-grained semantic correspondences through transformer-based masked prediction. The sew calibration mechanism imposes bidirectional constraints to compactly cluster same-identity features in a shared embedding space. Simultaneously, the masked modeling component acts as a cross-modal decoder that learns word-level representations, effectively discriminating subtle attribute differences. Importantly, our dual-encoder architecture strikes an optimal balance between representation expressiveness and computational efficiency by adopting a training-only decoder design. Extensive experiments on CUHK-PEDES, ICFG-PEDES, and RSTPReID datasets demonstrate that SCMM achieves state-of-the-art performance with Rank-1 accuracies of 73.81\%, 64.25\%, and 57.35\%, respectively. Thorough ablation studies confirm the efficacy of each proposed mechanism in establishing robust cross-modal patterns for multimodal perception and recognition.
\end{abstract}


\end{frontmatter}

% \linenumbers

\section{Introduction}
\label{sec:intro}
Cross-modal data retrieval and multimodal perception have emerged as fundamental research directions in pattern recognition, driven by the increasing demand to efficiently process and understand heterogeneous data sources \cite{yu2024discovering}. Person re-identification (Re-ID) exemplifies a challenging multimodal matching task that matches individuals across non-overlapping camera views by mining discriminative visual representations \cite{wu2025survey}. However, conventional Re-ID systems rely solely on visual queries, which inherently limits their applicability when visual references are unavailable in the database \cite{ding2025dynamic}. Text-Based Person Search (TBPS) overcomes this limitation by employing natural language descriptions as database queries, necessitating sophisticated knowledge engineering techniques to align visual and textual modalities in a shared semantic manifold \cite{feng2025homogeneous}. TBPS requires models to learn robust multimodal embeddings that capture both global semantic alignments and fine-grained data correspondences \cite{zhai2022lit}, thereby presenting unique challenges for representation learning and multimodal pattern analysis~\cite{gong2024crossmodal}. Edge-cloud collaborative intelligence further shows that multimodal representation learning is increasingly shaped by distributed deployment and model-optimization constraints~\cite{liu2026edgecloud}.

The effectiveness of these retrieval systems heavily depends on bridging the semantic gap between heterogeneous modalities while preserving discriminative features essential for accurate retrieval \cite{gnarnn}. Video anomaly detection studies similarly highlight the importance of reliable visual evidence in networked safety-critical perception systems~\cite{liu2025networking}. The primary challenge in TBPS arises from the semantic discontinuity between raw visual data and structured textual concepts, requiring features to be aligned across disparate embedding distributions~\cite{yu2025climbreid}. Large-scale clustering studies further highlight the strong value of globally organized representation spaces when retrieval systems must remain reliable at scale~\cite{ren2022global}. Recent advances in transformer-based architectures demonstrate that cross-modal attention mechanisms can facilitate rich feature interactions and semantic fusion between modalities \cite{qian2025survey}. Contemporary pattern recognition research highlights the importance of developing efficient alignment strategies that combine complementary information while ensuring computational tractability for real-time applications~\cite{bai2023textbased}. These advances are critical for cross-modal retrieval tasks that distinguish subtle visual differences guided by complex textual descriptions~\cite{caibc}.
Broader machine-learning studies also motivate the need for efficient, stable, and interpretable representation learning. GPU-accelerated optimal tree learning has clearly shown how scalable optimization can effectively improve discriminative decision structures on large datasets~\cite{ren2023gpuaccelerated}. In industrial monitoring, Cao et al.~\cite{cao2025comprehensive} made a valuable contribution to reliable soft sensing by emphasizing that accurate models should remain transparent under practical constraints.

\begin{figure}[!t]
  \centerline{\includegraphics[width=0.65\textwidth]{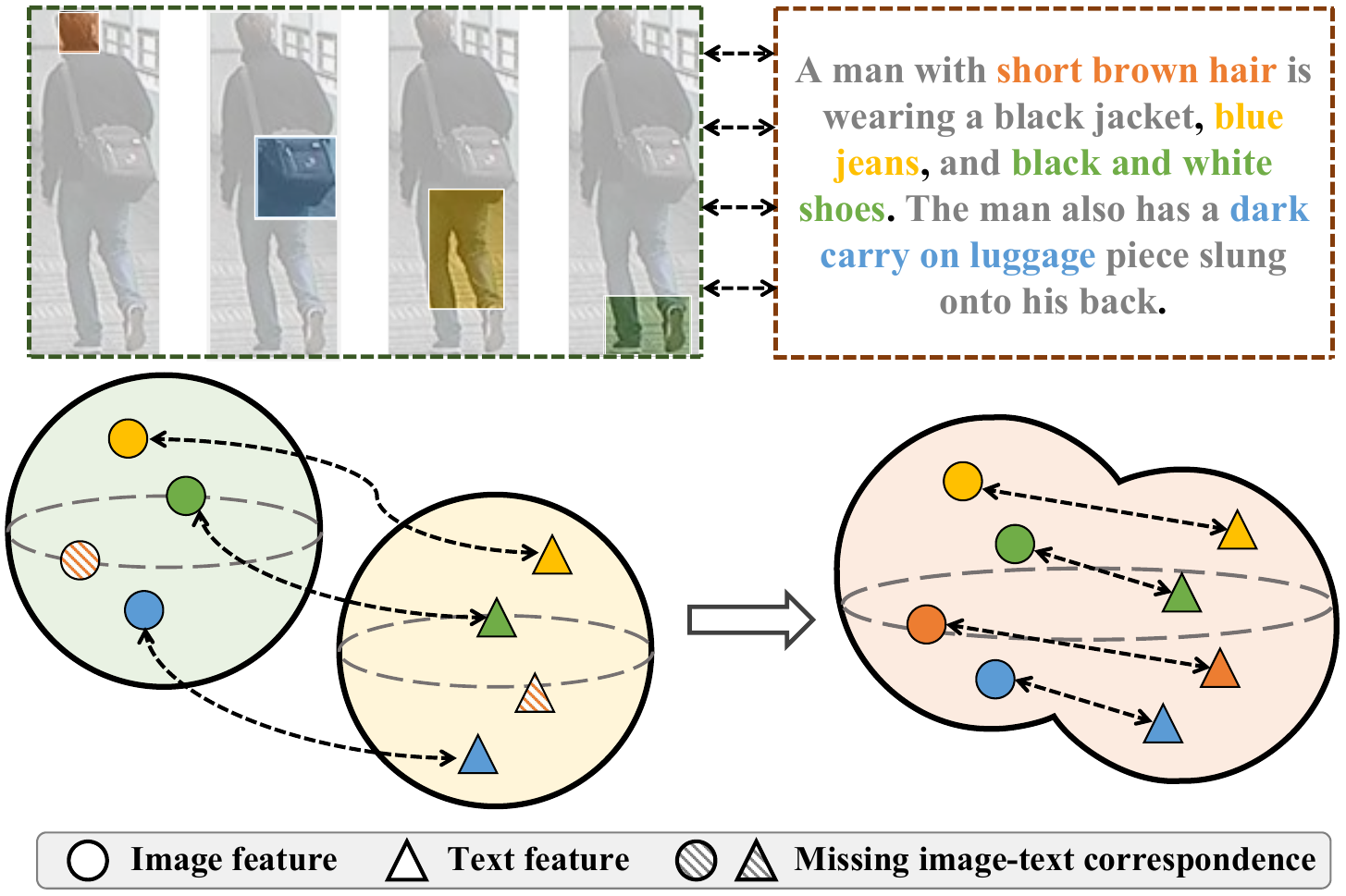}}
  \caption{Illustration of the motivation behind our method. In cross-modal tasks, a compact and well-aligned image-text feature distribution in the shared embedding space is crucial for bridging the inter-modal gap. Additionally, capturing fine-grained image-text correspondences is equally vital to distinguish between similar individuals for text-based person searching.}
  \label{fig:fig1}
\end{figure}

Compared to conventional Re-ID, the main objective of TBPS is to learn fine-grained cross-modal representations between visual and textual modalities~\cite{dssl}. As shown in Fig.~\ref{fig:fig1}, cross-modal representations have two key characteristics that contribute to effectively searching for a person from images: closely aligned cross-modal representations and fine-grained information correspondence. First, closely aligned cross-modal representations can effectively reduce the inter-modal gap, making it easier to locate specific persons using textual captions. Second, detailed correspondence is particularly important for TBPS, because fine-grained cross-modal information is often necessary to discriminate between similar persons~\cite{nafs}.

In the TBPS domain, numerous approaches have been proposed from two distinct perspectives. First, some methods \cite{dualpath, cmpmcmpc} employ only dual encoders and align the two modal representations using symmetric loss functions. Although computationally efficient, these methods suffer from limited feature alignment capabilities. Progressive knowledge distillation also suggests that staged supervision can stabilize representation transfer in compact learning pipelines~\cite{liu2026collaborative}. Alternative approaches \cite{alignbefore} utilize multi-modal models with transformer-based \cite{attention} cross-attention mechanisms to improve cross-modal feature alignment and interaction, a goal related to diffusion-guided semantic consistency under multimodal heterogeneity~\cite{liu2026diffusionguided}. However, such methods often require high computational costs due to the exhaustive fusion of all possible image-text pairs during inference, hindering their efficiency in real-time applications. Second, to obtain robust fine-grained features, recent TBPS works have designed multi-level \cite{vitaa, ssan, tipcb, safaNet}, multi-granularity \cite{mgel, ivt} matching strategies, and specialized attention modules \cite{gnarnn, cfine, acsa}. Semantic prototype learning offers a related view that class-level semantic anchors can improve recognition stability when distributions shift~\cite{liu2026semantic}. While these methods provide fine-grained features, they suffer from complex model architectures and expensive computations. Furthermore, the fine-grained features they produce are often limited and hinder significant performance improvements.

To address these persistent data matching challenges, we propose a novel framework, SCMM, structurally optimized for calibrating cross-modal representations in TBPS. Our approach relies on a dual-encoder architecture tailored for high-speed indexing and retrieval computationally fit for large-scale databases. To fully explore inter-modal semantics without sacrificing efficiency, we propose two learning objectives. First, a sew calibration loss steers alignment via quality-guided adaptive margins, selectively pushing apart negative instances and drawing in positive ones across modalities based on textual information density. Second, a Masked Caption Modeling (MCM) loss uncovers fine-grained word-level visual-textual connections. Operatively based on masked representation prediction, this phase activates a cross-modal decoder during training, thereby mapping fine-grained semantic connections without encumbering inference latency. Extensive experiments on three benchmarks clearly underscore our system's scalability and robustness, yielding state-of-the-art performance across all tested environments.
In summary, the primary contributions of this work are as follows:
\begin{itemize}
    \item We propose SCMM, an efficient dual-encoder framework augmented with an auxiliary cross-modal decoder, strategically designed to overcome the classical trade-off between computational scalability and multimodal alignment precision for text-based person search.
    \item We introduce a sew calibration loss anchored by quality-guided adaptive margins. By dynamically altering alignment boundaries based upon textual knowledge density, the loss diminishes the semantic gap and effectively organizes visual-textual knowledge representations.
    \item We formulate a masked caption modeling loss tailored to mine and formulate fine-grained cross-modal semantic correspondences, substantially heightening discriminatory capabilities among closely resembling instances.
    \item We validate the efficacy of SCMM through comprehensive experiments across CUHK-PEDES, ICFG-PEDES, and RSTPReID datasets, thereby validating the robustness of our method, with state-of-the-art Rank-1 accuracies corresponding to 73.81\%, 64.25\%, and 57.35\%.
\end{itemize}

The remainder of this paper is organized as follows. Sec.~\ref{sec:related} reviews related work on text-based person search and metric learning. Sec.~\ref{sec:methods} details the proposed SCMM framework, including the sew calibration and masked caption modeling losses. Sec.~\ref{sec:experi} presents the experimental setup, comprehensive results on three benchmarks, ablation studies, and qualitative visualizations. Finally, Sec.~\ref{sec:concl} concludes the paper with a summary of contributions and future research directions.

\section{Related Work}
\label{sec:related}
\subsection{Text-Based Person Search}
Text-based person search was first introduced by \cite{gnarnn}, which identifies person images in a gallery using only textual queries. Fundamentally, the task represents a critical application of multi-modal fusion, where the integration of heterogeneous data sources (visual and textual modalities) becomes essential for enhanced understanding and analysis capabilities. Multimodal medical surveys provide a related example in which heterogeneous visual and textual evidence must be fused carefully for dependable recognition~\cite{liu2025multimodala}. Early works utilized pre-processing methods to obtain external cues such as person segmentation \cite{vitaa} and human body landmarks \cite{verbalNet}. The evolution towards sophisticated fusion architectures has become crucial, as multi-view fusion combining complementary information from multiple sources shows significant promise for enhancing prediction accuracy and robustness in cross-modal tasks \cite{qian2025survey}.

In recent years, end-to-end frameworks based on attention mechanisms \cite{safaNet, ivt, acsa} have become prevalent, addressing the fundamental challenge of semantic mismatch between features of different modalities. Cross-modal attention is critical for performing image-text interaction, where advanced fusion strategies must consider the unique attributes of cross-modal matching tasks \cite{xie2025multiscale}. Existing methods can be broadly classified into attention-explicit and attention-implicit approaches. Attention-explicit methods \cite{gong2024crossmodal, acsa} design specific attention modules according to multi-granularity and multi-level strategies, implementing early fusion techniques that combine features at the input level. For example, NAFS \cite{nafs} conducts cross-modal alignments over full-scale features with a contextual non-local attention module, while CFine \cite{cfine} utilizes cross-attention for multi-grained global feature learning, achieving impressive results through knowledge transfer from the CLIP \cite{clip} model.
In contrast, attention-implicit methods \cite{safaNet, ivt} utilize transformer-based models with shared parameters to align cross-modal semantics implicitly, representing late fusion approaches that combine decisions from individual modality-specific models \cite{qian2025survey}. However, compared to performing in-depth cross-attention with task-driven approaches, existing methods face challenges in addressing information asymmetry problems and do not implement sufficient cross-interaction with generalized performance, highlighting the need for more sophisticated fusion techniques that enhance feature integration and establish deeper connections between global structure and local details \cite{xie2025multiscale}. To address these issues, various complementary strategies have emerged, such as serialized updating and matching \cite{sum}, implicit local alignment \cite{isa}, and cross-modal context sharing \cite{axm}. Furthermore, recent studies have explored robust feature learning under limited data constraints \cite{limit}, and introduced local-enhanced representations \cite{zhang2025localenhanced} along with comprehensive attribute prediction \cite{niu2024comprehensive} to capture fine-grained characteristics.
More recent transformer-based TBPS systems, including IRRA \cite{jiang2023crossmodal}, TP-TPS \cite{bai2023textbased}, and CLIP-enhanced variants such as CFine \cite{cfine}, have further strengthened retrieval performance through task-specific alignment designs or stronger pre-trained encoders. Their gains mainly come from richer feature interaction or better initialization. SCMM is positioned differently: it keeps the test-time dual-encoder retrieval path unchanged and sharpens supervision during training through quality-guided sew calibration and masked caption modeling. The resulting novelty lies in jointly improving person-specific global alignment and token-level correspondence without introducing a heavier inference-time fusion pipeline.

\subsection{Metric Learning}
Initially, metric learning used $L_2$ distance as the metric, with the goal of minimize the $L_2$ distance between samples of the same class. Some $L_2$-based metric learning methods include Siamese Networks and Triplet Networks \cite{facenet}. With the development of deep learning, researchers started using softmax-based loss functions in metric learning to learn a more discriminative distance metric. Additionally, increasing the margin between classes is an intuitive approach to learn a better metric space~\cite{xie2025multiscale}. L-Softmax introduced the concept of margin on the softmax function for the first time~\cite{verbalNet}. The widely used CosFace proposed large-margin cosine loss to learn highly discriminative deep features for face recognition~\cite{facenet}. Circle loss \cite{circle} proposed a simple loss based on a unified loss for metric learning and classification. Recently, some works introduced adaptive margin into marginal loss \cite{gong2024crossmodal}. They usually learn image quality implicitly and adjust the margin accordingly. In single-modal representation learning, they usually give large margins to high-quality samples for hard mining. Globally optimized clustering for K-center objectives by Ren et al.~\cite{ren2026global} provides a strong and valuable complementary view of how compact structure can be effectively imposed on large-scale representations. Compared to them, we try to solve a cross-modal matching problem where samples from two modalities have different information volume. We give greater tolerance to less informative samples in TBPS.
However, existing metric learning approaches face significant limitations when applied to cross-modal scenarios. First, traditional methods primarily focus on single-modal optimization and fail to adequately address the heterogeneity gap between different modalities \cite{facenet, circle}. Second, fixed margin constraints cannot adaptively handle the varying information density across modalities, leading to suboptimal alignment performance. Finally, most current approaches lack fine-grained correspondence modeling capabilities essential for distinguishing between similar cross-modal samples \cite{ding2025dynamic}.

\subsection{Masked Language Modeling}
MLM is a highly effective method for pre-training language models \cite{devlin2018bert} by randomly selecting a certain percentage of words from the input sentence and then predicting the masked words based on the context of other words. Many cross-modal pre-training models have utilized MLM in their methods. For example, the work in \cite{alignbefore} combines MLM with contrastive loss in the framework, which achieved impressive performance in their cross-modal tasks. The success of MLM in BERT \cite{lee2023lanobert} has proven its ability to adapt well to various downstream tasks, leading to generalized performance. In our fine-grained framework, the task-driven decoder utilizing masked caption modeling can facilitate generic cross-modal learning.
However, existing MLM-based cross-modal approaches face several limitations when applied to fine-grained person search tasks. First, traditional MLM methods primarily focus on intra-modal language understanding and lack explicit cross-modal interaction mechanisms, limiting their ability to establish detailed visual-textual correspondences \cite{zhai2022lit}. Second, most existing approaches apply MLM uniformly without considering the varying information density and quality of textual descriptions, leading to suboptimal learning efficiency. Automated soft-sensor design offers an important insight into deployment-facing model construction by suggesting a related principle: model selection and adaptation should respond to signal quality rather than remain fixed across all inputs~\cite{cao2025novel}. Survey work on diffusion-based anomaly detection further illustrates how generative objectives can strengthen discriminative learning when supervision is limited or imbalanced~\cite{liu2025anomaly,liu2025survey}. Finally, current methods often require complex multi-stage training procedures or heavy computational overhead during inference, hindering practical deployment in real-time applications \cite{mgel}.

\section{Proposed Methods}
\label{sec:methods}

Formally, given a set of images with corresponding captions, denoted as $X = \{(I_i, T_i)\}^N_{i=1}$, where each image $I_i$ and its description text $T_i$ is associated with a person ID $y_i$, the text-based person search task aims to retrieve the most relevant Rank $k$ (e.g., $k=1,5,10$) person images efficiently from a large-scale gallery using textual queries. To accomplish this, the task requires learning discriminative cross-modal representations that can bridge the semantic gap between visual and textual modalities while maintaining fine-grained correspondence for accurate retrieval. The fundamental challenge lies in designing neural network architectures that can align heterogeneous feature representations, where visual features capture spatial and appearance characteristics while textual descriptions encode semantic attributes and contextual information. Effective learning strategies must address both global alignment across modalities and local correspondences that enable discrimination between visually similar individuals.

\begin{figure}[!t]
  \centering
  \centerline{\includegraphics[width=\textwidth]{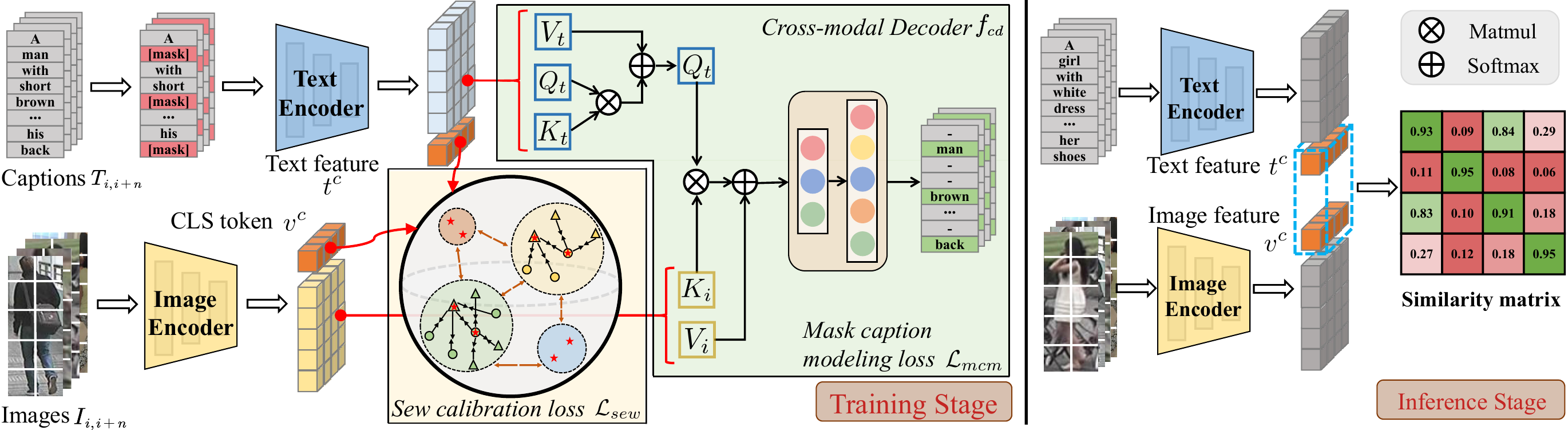}}
  \caption{Overview of our proposed SCMM. The framework consists of a dual-encoder for extracting image-text features and calibrating cross-modal representations with the sew calibration loss. We also include a decoder for performing cross-modal interaction with the task-driven Mask Caption Modeling. At the inference stage, we only utilize the classification (CLS) tokens from the dual-encoder to implement similarity search.}
  \label{fig:fig2}
  \vspace{-15px}
\end{figure}

To address these challenges in cross-modal learning, we propose SCMM, a streamlined yet effective framework as illustrated in Fig.~\ref{fig:fig2}. Our approach employs a dual-encoder architecture using Vision Transformer (ViT) \cite{vit} and BERT \cite{lee2023lanobert} as backbone networks for visual and textual feature extraction, respectively. The image encoder processes image patches from $I_i$ along with a vision classification (CLS) token, outputting an image feature sequence $\{v_i\}$ and a vision CLS token embedding $v_i^{c}$. Similarly, the text encoder generates a text feature sequence $\{t_i\}$ and a text CLS token embedding $t_i^{c}$ for caption $T_i$, following established practices in vision-language learning \cite{nafs, cfine}. Unlike CLIP-style pre-training \cite{clip}, which learns a generic joint space through symmetric image-text contrast over web-scale pairs, SCMM preserves a TBPS-oriented dual-encoder backbone and injects cross-modal interaction only in the training phase. This design keeps retrieval efficient at inference time while retaining token-level textual representations required by masked caption modeling. Our framework integrates two complementary learning mechanisms to calibrate cross-modal representations: the sew calibration loss with adaptive margin constraints for global alignment, and the masked caption modeling loss for establishing fine-grained cross-modal correspondences. The sew calibration objective also differs from the CLIP pre-training loss because it couples bidirectional matching and identity-classification constraints with sample-wise adaptive margins, thereby emphasizing person-specific alignment rather than generic image-text agreement. The following subsections elaborate on each component. We first introduce the sew calibration loss that dynamically aligns cross-modal features based on textual information quality. Subsequently, we present the masked caption modeling mechanism that facilitates fine-grained correspondence learning through transformer-based cross-modal attention and masked prediction. The total loss function combines these complementary objectives to jointly optimize global cross-modal alignment and fine-grained correspondence learning, enabling the framework to achieve superior performance while maintaining computational efficiency through the streamlined dual-encoder architecture.

\subsection{Sew Calibration Loss with Constraints}
\label{sec:scl}
\begin{figure}[!t]
  \centering
  \includegraphics[width=0.75\textwidth]{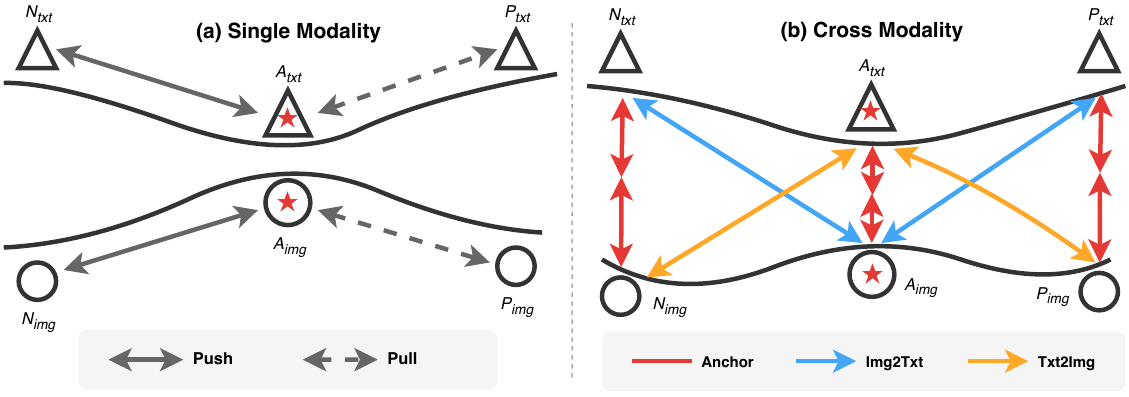}
  \caption{Illustration of sew calibration loss. The constraints are different between single-modal and cross-modal matching. $(A_{img}, A_{txt})$ denotes anchors in image-text feature distribution, while $(P_{img}, P_{txt})$ and $(N_{img}, N_{txt})$ denote positive and negative sample pairs, respectively. The sew calibration loss pushes negative sample pairs and pulls positive sample pairs, stitching cross-modal key information like a seam.}
  \label{fig:fig-method}
    \vspace{-15px}
\end{figure}

In the TBPS task, closely aligning cross-modal representations is crucial for effectively finding specific persons with textual captions. To address the heterogeneity between modalities, we propose a sew calibration loss that pushes each modality to a common space.
As shown in Fig.~\ref{fig:fig-method}(a), in single modality settings, a triplet distance constraint is widely utilized when embeddings are from a single modality distribution. In intra-class samples, the embeddings are pulled together, while inter-class samples are pushed away. In the cross-modal setting, we expect to impose constraints from both sides of the two embedding distributions. For example, in one-direction retrieval, we first need to align the features of a "perfect pair," i.e., $(I_i, T_i)$. The shortest distance from the image embedding ${v_i}$ to the text embedding should be ${t_i}$. Consequently, it is a perfect matching pair in person re-identification, and no other text feature will have a shorter distance (Eq.~\ref{eq:relaxed1}). We set the perfect pair as an image-text anchor $(A_{img}, A_{txt})$. Next, we impose another constraint between the image anchor $A_{img}$ and its corresponding positive text samples $P_{txt}(\mathds{1}(y_i = y_j, i \neq j))$ and negative text samples $N_{txt}$ (Eq. \ref{eq:relaxed2}). For the other direction, we put symmetric constraints on $A_{txt}$ and $P_{img}$, $N_{img}$. Fig.~\ref{fig:fig-method}(b) shows the proposed constraints for cross modality. Forces from both sides act like a seam to pull the two distributions together.

Take the image side as an example, the $L_2$ distance constraints are shown as follows:
\begin{eqnarray}
    D(A_{img}, A_{txt}) + \mathcal{M}_1< D(A_{img}, P_{txt}), \label{eq:relaxed1}\\
    D(A_{img}, P_{txt}) + \mathcal{M}_2 < D(A_{img}, N_{txt}), \label{eq:relaxed2}
\end{eqnarray}
where $D$ denotes $L_2$ distance. $\mathcal{M}_1$ and $\mathcal{M}_2$ are two margins, $\mathcal{M}_1 < \mathcal{M}_2$.  With a bi-directional margin, each modal feature of the same pedestrian target is compressed compactly, making decision boundaries clearer. We then relax the constraints to:
\begin{eqnarray}
    0.5D^2(A_{img}, A_{txt}) +\mathcal{M}_1< 0.5D^2(A_{img}, P_{txt}),\\
    0.5D^2(A_{img}, P_{txt})+ \mathcal{M}_2 < 0.5D^2(A_{img}, N_{txt}).
\end{eqnarray}
Subsequently, we change the above pairwise constraints into soft forms for better convergence following \cite{circle}.
\begin{eqnarray}
  \mathcal{L}^{Pull}_{match} = log\big[1+\sum\limits^K_{k=1}exp(\alpha(\bar{v}_i^{c}\bar{t}_k^{c} - \bar{v}_i^{c} \bar{t}_i^{c} + \mathcal{M}_1)) \big],
\end{eqnarray}

\begin{eqnarray}
   \mathcal{L}^{Push}_{match} = log\big[1+\sum\limits^K_{k=1}\sum\limits^J_{j=1}exp(\alpha(\bar{v}_i^{c}\bar{t}_j^{c} - \bar{v}_i^{c}\bar{t}_k^{c} + \mathcal{M}_2)) \big],
\end{eqnarray}
where $\bar{v}^{c}$ and $\bar{t}^{c}$ are CLS token features after normalizing, $K$ and $J$ denote the number of positive and negative samples in this batch, respectively. $\alpha$ is a scale parameter. As a result, the image-to-text matching part of the sew calibration loss is formulated as below:
\begin{eqnarray}
\mathcal{L}^{I2T}_{match} =\mathcal{L}^{Pull}_{match} +\mathcal{L}^{Push}_{match}.
\end{eqnarray}
The constraints in Eq.~\ref{eq:relaxed2} can also be used to impose a classification loss in a similar way. As there is no difference for perfect positive samples in the classification task, we omit to constrain Eq.~\ref{eq:relaxed1}. Formally, the loss for our person ID classification part is as follows \cite{cmpmcmpc}:
\begin{gather}
\begin{align}
\mathcal{L}^{I2T}_{id} &= \frac{1}{n}\sum\limits_{i=1}^n-log\big(\frac{e^{(\alpha(s_{y_i,i}-\mathcal{M}_2))}}{e^{(\alpha(s_{y_i,i}-\mathcal{M}_2))}+\sum\limits_{j\neq{y_i}}e^{(\alpha(s_{y_j,i}))}}),
\end{align}\\
s_{i,j} = \omega^T_i \hat{v}_j, \quad
\hat{v}_i = (v^{c}_i)^T \bar{t}^{c}_i \cdot \bar{t}^{c}_i,
\end{gather}
where $n$ is batch size, and $\omega$ represents the classification weight after normalization. $\hat{v}_i$ can be explained as the projection of image representation $v^{cls}_i$ onto the normalized text representation $\bar{t}^{c}_i$. 
Intuitively, $\hat{v}_i$ keeps only the component of the image CLS representation that lies along the paired text direction. The ID classifier therefore scores whether the image embedding is projected onto the correct text-conditioned identity axis, which makes Eqs.~(8) and (9) easier to interpret: the margin term encourages the projected image feature to stay closer to its matched identity than to competing identities in the shared space.

$\mathcal{L}^{T2I}_{match}$ and $\mathcal{L}^{T2I}_{id}$ are in the same form as above, but the change is focused on the text-to-image. Both matching and classification loss have identical decision boundaries. Equipped with our proposed cross-modal constraints, the sew calibration loss can effectively reduce the gap between image and text feature distributions. Although the margin restrictions allow our model to learn better cross-modal representations, using a fixed margin $\mathcal{M}$ in all cases may not be flexible enough. A large margin constraint makes model learning difficult, while too small a margin does not impose a significant constraint. An adaptive margin guided by quality can be more effective. In TBPS, the texts come from the annotator's descriptions of the images. Images are expected to have complete information, while texts have varying amounts of information. Thus, we adjust the margin value based on the quality of the text description. Token length is used here as a coarse proxy for textual information density rather than a direct measure of semantic usefulness: a short caption may still contain highly discriminative attributes, whereas a longer description can include redundancy or exceed the text encoder's effective context usage. The adaptive margin therefore serves as a monotonic prior on annotation richness, giving looser constraints to less detailed captions while preserving a simple formulation that is stable across datasets. Based on this design, we compute the adaptive margin for each image-text pair according to its text total tokens length $\mathcal{T}_i$:
\begin{eqnarray}
\mathcal{M}_{i} = \mathcal{M}_{min} + \frac{(\mathcal{M}_{max}-\mathcal{M}_{min})\cdot (\mathcal{T}_i-\mathcal{T}_{min})}{\mathcal{T}_{max}-\mathcal{T}_{min}},
\end{eqnarray}
where $\mathcal{M}_{max}$ and $\mathcal{M}_{min}$ are upper and lower bounds of margins. $\mathcal{T}_{max}$ and $\mathcal{T}_{min}$ are bounds of the captions length, respectively. We set $\mathcal{T}_{max}$ and $\mathcal{T}_{min}$ according to the different dataset captions length distributions. After that, we utilize $\mathcal{M}_{i}$ to replace the fixed margin $\mathcal{M}$ above. We simply set $\mathcal{M}_1 = \mathcal{M}_2=\mathcal{M}_{i}$.
Each $\mathcal{M}_i$ is computed independently from the token count of caption $T_i$ before the batch losses are aggregated. The normalization uses the dataset-level bounds $(\mathcal{T}_{min},\mathcal{T}_{max})$ rather than the current mini-batch statistics, so the adaptive margin is not re-scaled by the composition of a particular batch.

\subsection{Masked Caption Modeling Loss}
TBPS is a fine-grained cross-modal task, which means that only caption-level discrimination is not enough. If the textual captions of two persons differ in a few words, a TBPS method can not retrieve a specific person without word-level discrimination. Although there are many works to establish word-level discrimination capacities, such methods are complex and limited, hindering performance boost. To solve this issue, we propose masked caption modeling to establish detailed image-text relationships. Furthermore, by utilizing MCM, our framework can perform more generic cross-modal learning.

Inspired by \cite{zhai2022lit, devlin2018bert}, we add a masked prediction task on the text branch. Concretely, we mask a portion of text tokens (replacing them with a learnable token vector) and employ a cross-modal decoder $f_{cd}$. The text encoder outputs the corresponding text features, and the decoder maximizes the conditional likelihood of the masked text feature $t_n$ given the image feature sequence $\{v_i\}$ and text feature sequence $\{t_i\}$.

\begin{eqnarray}
\mathcal{L}_{mcm}=-\sum_{n=1}^{N} \text{log } P_{\theta}(t_n|\{t_i\}, \{v_i\}),
\end{eqnarray}
where $N$ is the total masked tokens numbers in a caption.
 
\begin{algorithm}[!t]
\label{alg:scmm}
\begin{minipage}{0.98\linewidth}
\begingroup
\setlength{\parindent}{0pt}
\newcommand{\algindent}{\hspace*{1.5em}}
\newcommand{\algindenttwo}{\hspace*{3.0em}}
\textbf{Input:} Image $I$ and text $T$; a batch of $n$ paired samples $\mathbb{B}=\{(I^1,T^1),(I^2,T^2),\cdots,(I^n,T^n)\}$\par
\textbf{Output:} Training loss $(\mathcal{L}_{sew}, \mathcal{L}_{mcm})$ or embeddings $(I_{cls}, T_{cls})$\par
\textbf{for each} $T^i$ in $\mathbb{B}$ \textbf{do}\par
\algindent $T^i \leftarrow \text{mask}(T^i)$\par
\textbf{end for}\par
\textbf{for} $i \leftarrow 1$ to $n$ \textbf{do}\par
\algindent \alghl{$(I^i_{cls}, I^i_1, \cdots, I^i_c) \leftarrow ViT(I^i)$}\par
\algindent \alghl{$(T^i_{cls}, T^i_1, \cdots, T^i_c) \leftarrow Bert(T^i)$}\par
\algindent \textbf{if} Training Stage \textbf{then}\par
\algindenttwo \alghl{$(I^i_{cls}, T^i_{cls}) \leftarrow BatchNorm(I^i_{cls}, T^i_{cls})$}\par
\algindenttwo \alghl{$I^i_{context}, T^i_{context} \leftarrow (I^i_1, \cdots, I^i_c), (T^i_1, \cdots, T^i_c)$}\par
\algindenttwo \alghl{$T^i_{context} \leftarrow Attn(T^i_{context})$}\par
\algindenttwo \alghl{$T^i_{cross} \leftarrow CrossAttn(T^i_{context}, I^i_{context})$}\par
\algindenttwo \alghl{$\mathcal{L}_{sew}^i \leftarrow SewCalibration(I^i_{cls}, T^i_{cls})$}\par
\algindenttwo \alghl{$\mathcal{L}_{mcm}^i \leftarrow MCM(T^i_{cross}, T^i)$}\par
\algindent \textbf{else}\par
\algindenttwo $\{(I^1_{cls}, T^1_{cls}), (I^2_{cls}, T^2_{cls}), \cdots, (I^i_{cls}, T^i_{cls})\} \leftarrow (I_{cls}, T_{cls})$\par
\algindent \textbf{end if}\par
\textbf{end for}\par
\textbf{if} Training Stage \textbf{then}\par
\algindent \textbf{return} $(\mathcal{L}_{sew}, \mathcal{L}_{mcm})$\par
\textbf{else}\par
\algindent \textbf{return} $(I_{cls}, T_{cls})$\par
\textbf{end if}\par
\endgroup

\end{minipage}
\caption{Training Procedure of SCMM}
\end{algorithm}
\FloatBarrier

As shown in Fig.~\ref{fig:fig2}, the cross-modal decoder part, the decoder $f_{cd}$ takes both unmasked text tokens and masked tokens in their original order as the input. The multi-head self-attention \cite{attention} first encodes the text features as $Q_t, K_t, V_t$, while the cross self-attention further improves the text features by taking into account the encoded image features as $K_i, V_i$ for visual context. The final linear projection layer has the same number of output channels as the text vocabulary and computes the cross-entropy loss between the reconstructed and original words only on masked text tokens. It should be noted that $f_{cd}$ is only used during training and not during inference.

\subsection{Total Loss}
The total loss function of SCMM integrates our two complementary learning mechanisms to achieve comprehensive cross-modal alignment and correspondence learning. Our unified optimization objective combines the sew calibration loss and the masked caption modeling loss to jointly address both global alignment and fine-grained correspondence challenges in cross-modal representation learning. The sew calibration loss $\mathcal{L}_{sew}$ includes bidirectional matching and classification components that ensure symmetric cross-modal alignment:
\begin{eqnarray}
\mathcal{L}_{sew} = 
    \mathcal{L}_{match}^{I2T} + \mathcal{L}_{match}^{T2I} +
    \mathcal{L}_{id}^{I2T} + \mathcal{L}_{id}^{T2I},
\end{eqnarray}
where the image-to-text and text-to-image matching losses establish cross-modal correspondence constraints, while the identification losses enforce discriminative learning for person-specific features. Importantly, the bidirectional formulation ensures that both modalities are effectively aligned in the shared embedding space through symmetric constraints. Subsequently, the complete loss function combines both components through weighted summation:
\begin{eqnarray}
\mathcal{L} = \lambda_1 \mathcal{L}_{sew} + \lambda_2 \mathcal{L}_{mcm},
\end{eqnarray}
where $\lambda_1$ and $\lambda_2$ are hyperparameters balancing global alignment and fine-grained correspondence learning. Through this joint optimization, the framework simultaneously learns compact cross-modal representations while capturing detailed visual-textual relationships for discriminating similar individuals. The complementary loss components create synergistic effects ensuring comprehensive representation learning, addressing both inter-modal gap reduction and fine-grained correspondence modeling. Regarding computational complexity, SCMM maintains efficient processing through its streamlined dual-encoder architecture.

\section{Experiments}
\label{sec:experi}

\subsection{Datasets and Evaluation Metric}
We evaluate SCMM on three benchmarks: CUHK-PEDES \cite{gnarnn}, ICFG-PEDES \cite{ssan}, and RSTPReid \cite{dssl}. As the first large-scale benchmark for text-based person search tasks, CUHK-PEDES contains 40,206 images of 13,003 person IDs collected from five person re-identification datasets, with each image having two different textual captions with an extensive vocabulary and an average sentence length of 23.5 words. Additionally, the testing set comprises 3,074 images and 6,148 descriptions of 1,000 persons. ICFG-PEDES contains 54,522 images of 4,102 persons, where each image's corresponding description has an average length of 37 words, and the testing set consists of 19,848 image-text pairs of 1,000 persons. RSTPReID contains 20,505 images of 4,101 persons, with each pedestrian having five images. It is divided into a training set with 3,701 persons, a validation set with 200 persons, and a testing set with 200 persons. For our evaluation metric, we report the Rank $k$ ($k$=1, 5, 10) text-to-image accuracy, which is commonly used in previous works to evaluate text-based person search. Given a textual description as the query, if the top-$k$ retrieved images contain any person corresponding to the query, we consider it a successful person search.

\subsection{Implementation Details}
In our visual-textual dual-encoder, we extract visual representations using the ViT-Base pre-trained on ImageNet \cite{imagenet}. The images are resized to 224 $\times$ 224 pixels. For textual representations, we use the BERT-Base-Uncased model pre-trained on the Toronto Book Corpus and Wikipedia. The representation dimension is set to 768, and the feature sequence lengths are set to 197 and 100, respectively.
During the training phase, we use a batch size of 64 and train for 60 epochs. We use Adam optimizer with an initial learning rate of 0.001. To augment our data, we apply a random horizontal flipping operation, and we use a mask ratio of 0.1 for randomly masking text tokens.
After selecting this value on CUHK-PEDES through the ablation in Table~\ref{tab:mr}, we keep the same 0.1 mask ratio for ICFG-PEDES and RSTPReID rather than tuning a separate ratio for each dataset. Only the caption-length bounds $(\mathcal{T}_{min},\mathcal{T}_{max})$ are adjusted to match dataset-specific caption distributions.
The minimum and maximum textual information length boundaries $\mathcal{T}_{min}$ and $\mathcal{T}_{max}$ are set to 20-60, 25-65, and 22-60 according to the caption length distributions of CUHK-PEDES, ICFG-PEDES, and RSTPReID, respectively. The bounds of the upper and lower margin $\mathcal{M}$ are set to 0.4 and 0.6, and the scale parameter $\alpha$ is set to 32. For each loss in the total loss function, the balance factors $\lambda_1$ and $\lambda_2$ are set equal to 1. During the testing phase, we apply a re-ranking post-processing approach to improve search performance following NAFS~\cite{nafs}. We conduct the experiments on four NVIDIA 2080Ti with PyTorch.

\subsection{Comparison with State-of-the-art Methods}
\begin{table}[!t]
    \centering
    \caption{\label{tab:e1}Comparison with SOTA methods on the CUHK-PEDES dataset. Rank1, Rank5, and Rank10 accuracies (\%) are reported. Bold number represents the best score. R and C denote the re-ranking post-processing operations and image encoder pre-trained on CLIP model.}
    \setlength{\tabcolsep}{4.8mm}{
    \resizebox{.75\textwidth}{!}{
    \begin{tabular}{c|ccc}
        \toprule
        \textbf{Method} & \textbf{Rank1} & \textbf{Rank5} & \textbf{Rank10}\\
        \hline
        GNA-RNN \cite{gnarnn} & 19.05    & -   & 53.64 \\
        CMPM/C \cite{cmpmcmpc}  & 49.37 & - & 79.27\\
        Dual-path \cite{dualpath}   & 44.40 & 66.26 & 75.07\\
        ViTAA \cite{vitaa}  & 55.97 & 75.84 & 83.52\\
        VP Net \cite{verbalNet}  & 58.83 & 81.25 & 86.72\\
        SUM \cite{sum}  & 59.22 & 80.35 & 87.60\\
        CLIP \cite{clip} & 60.67    & 81.99   & 88.87 \\
        DSSL \cite{dssl}  & 59.98 & 80.41 & 87.56\\ 
        MGEL \cite{mgel}  & 60.27 & 80.01 & 86.74\\
        SSAN \cite{ssan} & 61.37 & 80.15 & 86.73\\
        TestReID \cite{limit}  & 64.08 & 81.73 & 88.19\\
        ACSA \cite{acsa} & 63.56 & 81.40 & 87.70\\
        IVT \cite{vit}  & 64.00 & 82.72 & 88.95\\
        SAFA Net \cite{safaNet}  & 64.13 & 82.62 & 88.40\\
        TIPCB \cite{tipcb}  & 64.26 & 83.19 & 89.10\\
        CAIBC \cite{caibc} & 64.43 & 82.87 & 88.37\\
        AXM Net \cite{axm}  & 64.44 & 80.52 & 86.77\\
        CFine \cite{cfine} & 65.07 & 83.01 & 89.00\\
        ISANet \cite{isa}  & 63.92 & 82.15 & 87.69\\
        CAPL~\cite{niu2024comprehensive} & 65.63 & 84.81 & 90.21\\
        LERF~\cite{zhang2025localenhanced} & 65.84 & 84.24 & 90.22\\
        \hline
        \textbf{SCMM} & 67.71 & 84.57 & 89.44\\
        \textbf{SCMM+R} & 71.09 & 86.78 & 91.23\\
        \textbf{SCMM+C} & 69.61 & 86.01 & 90.90\\
        \hline
        \rowcolor{lightkeycolor}
        \textbf{SCMM+R+C} & \textbf{73.81} & \textbf{88.89} & \textbf{92.77}\\
        \bottomrule
    \end{tabular}}}
    \vspace{-5px}
\end{table}

\subsubsection{Results on CUHK-PEDES.}
Table~\ref{tab:e1} compares our framework and previous methods on CUHK-PEDES. To keep the comparison fair, it is useful to read the results in three groups: standard methods without re-ranking or CLIP-enhanced initialization, methods with re-ranking, and variants that additionally import a CLIP-pre-trained visual encoder. Under the standard setting, our proposed SCMM outperforms the previous best method CFine \cite{cfine} by 2.64\%, 1.56\%, and 0.44\% on Rank1, Rank5, and Rank10, respectively. The state-of-the-art results on CUHK-PEDES show the effectiveness of SCMM in the same inference regime. Re-ranking then provides an additional boost, raising Rank1 to 71.09\%. Furthermore, with a stronger image encoder pre-training on CLIP, SCMM achieves 73.81\%, 88.89\%, and 92.77\% on the three metrics, respectively. In this comparison, SCMM+C replaces only the visual backbone with CLIP-pre-trained initialization \cite{clip} while retaining the BERT-based text encoder \cite{lee2023lanobert}. That configuration isolates the effect of transferring a stronger visual encoder into the proposed framework and keeps the masked caption modeling branch compatible with token-level word prediction. Replacing the text encoder with a CLIP text tower is also meaningful, but it would require a joint redesign of the textual token interface and the training-only decoder rather than an isolated backbone swap. Overall, the grouped comparison shows that SCMM scales consistently across the base setting, stronger visual pre-training, and extra post-processing operations.

\begin{table}[!t]
    \vspace{-10px}
    \centering
    \caption{\label{tab:e2}{Quantitative results on the ICFG-PEDES dataset.}}
    \setlength{\tabcolsep}{4.8mm}{
    \resizebox{.75\textwidth}{!}{
    \begin{tabular}{c|ccc}
        \toprule
        \rowcolor{gray!8}
        \textbf{Method} & \textbf{Rank1} & \textbf{Rank5} & \textbf{Rank10}\\
        \hline
        \hline
        CMPM/C \cite{cmpmcmpc}  & 43.51 & 65.44 & 74.26\\
        Dual-path \cite{dualpath}   & 38.99 & 59.44 & 68.41\\
        ViTAA \cite{vitaa}  & 50.98 & 68.79 & 75.78\\
        SSAN \cite{ssan} & 54.23 & 72.63 & 79.53\\
        CLIP \cite{clip} & 53.96 &73.69&80.43 \\
        TIPCB \cite{tipcb}  & 54.96 & 74.72 & 81.89\\
        IVT \cite{ivt}  & 56.04 & 73.60 & 80.22\\
        CFine \cite{cfine} & 55.69 & 72.72 & 79.46\\
        CFine+C \cite{cfine} & 60.83 & 76.55 & 82.42\\
        IRRA \cite{jiang2023crossmodal} & 63.46 & 80.25 & 85.82 \\
        TP-TPS \cite{bai2023textbased} & 60.64 & 75.97 & 81.76 \\
        ISANet \cite{isa}  & 57.73 & 75.42 & 81.72\\
        LERF~\cite{zhang2025localenhanced} & 57.23 & 76.64 & 83.11\\
        CANC \cite{gong2024crossmodal}  & 60.52 & 78.36 & 84.13\\
        \hline
        \textbf{SCMM} & 60.20 & 75.97 & 81.78\\
        \textbf{SCMM+R} & 62.17 & 85.74 & 89.67\\
        \textbf{SCMM+C} & 62.29 & 77.15 & 82.52\\
        \hline
        \rowcolor{lightkeycolor}
        \textbf{SCMM+R+C} & \textbf{64.25} & \textbf{86.95} & \textbf{90.70}\\
        \bottomrule
    \end{tabular}}}
    \vspace{-10px}
\end{table}

\begin{table}[!t]
    \centering
    \caption{\label{tab:e3}{Quantitative results on the RSTPReid dataset.}}
    \setlength{\tabcolsep}{5.5mm}{
    \resizebox{.75\textwidth}{!}{
    \begin{tabular}{c|ccc}
        \toprule
        \textbf{Method} & \textbf{Rank1} & \textbf{Rank5} & \textbf{Rank10}\\
        \hline
        CLIP \cite{clip} & 50.10    & 76.10  & 84.95 \\
        DSSL \cite{dssl}  & 32.43 & 55.08 & 63.19\\
        SUM \cite{sum}  & 41.38 & 67.48 & 76.48\\
        SSAN \cite{ssan} & 43.50 & 67.80 & 77.15\\
        IVT \cite{ivt}  & 46.70 & 70.00 & 78.80\\
        ACSA \cite{acsa} & 48.40 & 71.85 & 81.45\\  
        LERF~\cite{zhang2025localenhanced} & 46.75 & 71.30 & 81.60\\
        CFine \cite{cfine} & 45.85 & 70.30 & 78.40\\
        CFine+C \cite{cfine} & 50.55 & 72.50 & 81.60\\
        TP-TPS \cite{bai2023textbased} & 50.65 & 72.45 & 81.20\\
        \hline
        \textbf{SCMM} & 50.75 & 74.20 & 81.70\\
        \textbf{SCMM+R} & 55.35 & 77.30 & 84.25\\
        \textbf{SCMM+C} & 51.95 & 73.50 & 82.45\\
        \hline
        \rowcolor{lightkeycolor}
        \textbf{SCMM+R+C} & \textbf{57.35} & \textbf{77.50} & \textbf{85.50}\\
        \bottomrule
    \end{tabular}}}
    \vspace{-5px}
\end{table}

\subsubsection{Results on ICFG-PEDES and RSTPReid.} 
We also utilized other benchmarks to validate SCMM's performance and generalization. The ICFG-PEDES and RSTPReid datasets are more challenging compared to CUHK-PEDES, and our method significantly outperformed all state-of-the-art methods on these two datasets by a large margin, as reported in Tables~\ref{tab:e2} and~\ref{tab:e3}. Compared with the state-of-the-art results \cite{cfine} on ICFG-PEDES, SCMM achieved significant improvements, with scores of 60.20\% \textbf{(+4.51\%)}, 75.97\% \textbf{(+3.25\%)}, and 81.78\% \textbf{(+2.32\%)} on Rank1, Rank5, and Rank10, respectively. On the RSTPReid dataset, we achieved scores of 50.75\% \textbf{(+4.90\%)}, 74.20\% \textbf{(+3.90\%)}, and 81.70\% \textbf{(+3.30\%)} on the three metrics. With re-ranking post-processing, we were able to achieve scores of 62.17\% and 55.35\% on Rank1 for the two benchmarks, respectively.
The gain on RSTPReid is more modest before re-ranking: SCMM improves Rank1 from 50.65\% with TP-TPS \cite{bai2023textbased} to 50.75\%, which is still a positive but narrower margin than the headline results obtained with re-ranking or CLIP-enhanced initialization. A plausible reason is that RSTPReid contains fewer testing identities and many visually similar pedestrians \cite{dssl}, making fine-grained distinctions harder when retrieval relies only on the base embedding geometry. Re-ranking should therefore be interpreted as a separate post-processing enhancement rather than part of the core base-model comparison, and it brings a significant boost because the learned embeddings already preserve compact cross-modal neighbor relations. Moreover, by utilizing the CLIP pre-trained image model as an encoder, we achieved even better results, with scores of 64.25\%, 86.95\%, and 90.70\% on ICFG-PEDES, and scores of 57.35\%, 77.50\%, and 85.50\% on RSTPReid. Consequently, the two challenging benchmark results demonstrate the robustness and scalability of SCMM across the base setting and the stronger CLIP-initialized variant.

\subsection{Ablation Studies}
\begin{table}[!t]
    \centering
    \caption{\label{tab:e4}{Performance comparisons of various components.}}
    \setlength{\tabcolsep}{1.5mm}{
    \resizebox{.90\textwidth}{!}{
    \begin{tabular}{l|ccccc|ccc}
        \toprule
        \rowcolor{gray!8}
        \textbf{ID} & Res & ViT & Sew-F & Sew-A & MCM & \textbf{Rank1} & \textbf{Rank5} & \textbf{Rank10}\\
        \hline
         \textbf{1} & \Checkmark &  &  &  &  & 60.39 & 80.22 & 86.53\\
         \textbf{2} &  & \Checkmark &  &  &  & 62.29 & 81.25 & 87.22\\
         \hline
         \textbf{3} &  & \Checkmark & \Checkmark &  &  & 65.41 & 82.56 & 88.01\\
         \textbf{4} &  & \Checkmark &  & \Checkmark &  & 66.02 & 83.33 & 88.61\\
         \hline
         \textbf{5} &  & \Checkmark &  &  & \Checkmark & 65.16 & 81.95 & 87.93\\
         \hline
         \rowcolor{lightkeycolor}
         \textbf{6} &  & \Checkmark &  & \Checkmark & \Checkmark & \textbf{67.71} & \textbf{84.57} & \textbf{89.44}\\
        \bottomrule
    \end{tabular}}}
     \fontsize{8.3pt}{2pt}\selectfont {
  \begin{minipage}{\columnwidth}
    \vspace{1px}
    \textit{Notes}: {Res and ViT denote ResNet and Vision Transformer as image encoders, respectively. Sew-F and Sew-A refer to the fixed and adaptive margin versions of the sew calibration loss, respectively. Model 2 is our baseline.}
  \end{minipage}
  }
    \vspace{-10px}
\end{table}

\subsubsection{Analysis of Model Components}
To validate SCMM's components, we evaluate their contributions on CUHK-PEDES (Table~\ref{tab:e4}). \textbf{Models 1} and \textbf{2} compare ResNet-50 and ViT-Base encoders, with ViT showing superior performance (+1.90\% Rank1). Using Model 2 as the baseline, \textbf{Models 3} and \textbf{4} demonstrate that the sew calibration loss improves alignment, with adaptive margins (\textbf{Model 4}) outperforming fixed margins (\textbf{Model 3}). \textbf{Model 5} shows that MCM significantly boosts performance by mining fine-grained details. Notably, the full SCMM framework (\textbf{Model 6}) combines these components to achieve the best results, outperforming the baseline by 5.42\% in Rank1 accuracy. The separate gains from Sew-A (+3.73\%) and MCM (+2.87\%) are therefore not strictly additive, indicating partial overlap in the errors they correct. At the same time, the full model still surpasses either single component by 1.69\%--2.55\%, which suggests a positive interaction: sew calibration improves the global embedding geometry, while MCM contributes complementary token-level cues that are more useful once the two modalities are already better aligned. The consistent improvement confirms that jointly optimizing global alignment and fine-grained correspondence is essential for effective cross-modal representation learning.
\begin{figure}[!t]
  \centerline{\includegraphics[width=\textwidth]{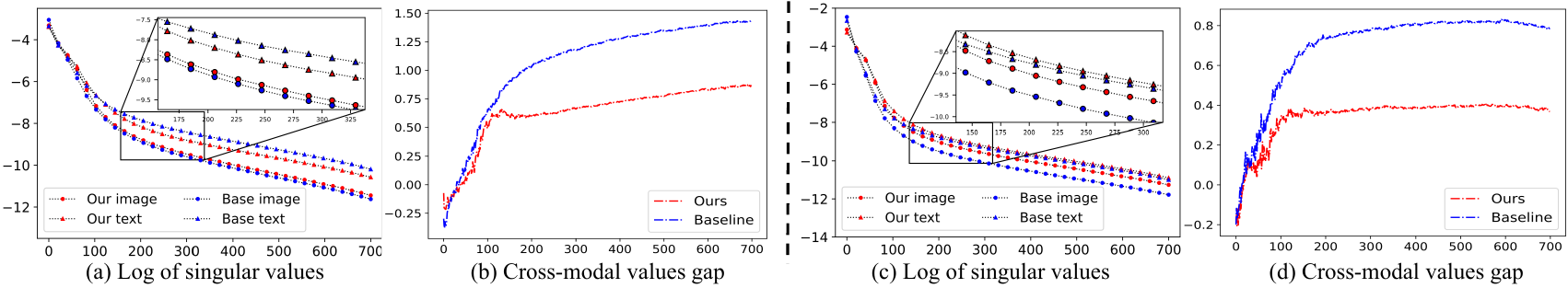}}
  \caption{Comparison of singular values for image-text embedding features across CUHK-PEDES (a-b) and ICFG-PEDES (c-d) datasets. (a) and (c) depict the distribution of singular values, where a smaller inter-line gap indicates a closer cross-modal distribution. (b) and (d) present the logarithmic difference in singular values between the baseline and our approach for CUHK-PEDES and ICFG-PEDES, respectively.}
  \label{fig:fig3}
\end{figure} 

\subsubsection{Impact of Sew Calibration Loss for Reducing Cross-modal Gap}
Benefiting from the sew calibration loss, which reduces the cross-modal gap, the learned representation distributions are closer than with the basic loss. To demonstrate this, we illustrate a comparison of singular value decomposition on cross-modal representations \cite{xie2025multiscale}. Specifically, Fig.~\ref{fig:fig3}(a) and (c) present the baseline and SCMM's singular value decomposition for the text and image modalities. We computed the distance between these two modalities at specific singular values and reflected them in Fig.~\ref{fig:fig3}(b) and (d). Intuitively, the smaller the distance between different modality features, the closer the distribution learned by the model. We find that the sew calibration loss performs better representation distributions in the common embedding space. It ensures closer cross-modal representation distributions and a smaller inter-modal gap than the baseline.

\subsubsection{Impact of Masked Caption Modeling}
The masked caption modeling operation in fine-grained cross-modal interaction is critical in our framework. It comprises token masking, an attention module, and masked tokens prediction. As shown in Table~\ref{tab:e5}, we explore the effectiveness of MCM components on the CUHK-PEDES. First, only masking on the input text tokens without a reconstruction task behaves like a random erase text augmentation. Remarkably, the augmentation can already bring +1.36\% marginal improvement and reach 63.65\% on Rank1. It shows that the details provided by word tokens are helpful for fine-grained recognition.
Next, if we only utilize the attention mechanism to enhance cross-modal representations without the mask caption modeling, it can achieve 63.83\% on Rank1. We explain this improvement as the cross-attention brings details from sequence tokens to the CLS token. Meanwhile, image features also contribute a lot to this process. 

On the other hand, we also design an experiment of mask caption modeling without the cross attention. The decoder $f_{cd}$ directly predicts all masked tokens from dual-encoder outputs. In this experiment, we observe 64.28\% on Rank1. It proves the reconstruction tasks in the decoder help guide the text encoder to learn richer and more refined representations. We also notice that with all three components, the MCM achieves 65.16\% on Rank1. With the help of cross attention from image features, the CLS token ensembles all those rich information. 

\begin{table}[!t]
    \centering
    \caption{\label{tab:e5}{MCM components on the CUHK-PEDES dataset.}}
    \setlength{\tabcolsep}{1.5mm}{
    \resizebox{.80\textwidth}{!}{
    \begin{tabular}{ccc|ccc}
        \toprule
        \rowcolor{gray!8}
         Mask & Attention & Caption Modeling & \textbf{Rank1} & \textbf{Rank5} & \textbf{Rank10}\\
        \hline
        \Checkmark &  &  & 63.65 & 81.84 & 87.51\\
         & \Checkmark &  & 63.83 & 80.87 & 86.35\\
        \Checkmark &  & \Checkmark & 64.28 & 81.97 & 87.75\\
        \hline
        \rowcolor{lightkeycolor}
        \Checkmark & \Checkmark & \Checkmark & \textbf{65.16} & \textbf{81.95} & \textbf{87.93}\\
        \bottomrule
    \end{tabular}}}
    \vspace{-10px}
\end{table}
\subsubsection{Analysis of Generalisability Validation}

\begin{table}[!t]
    \centering
    \caption{\label{tab:e6}{Performance analysis for cross-domain validation.}}
    \setlength{\tabcolsep}{4.5mm}{
    \resizebox{.75\textwidth}{!}{
    \begin{tabular}{c|ccc}
        \toprule
        \rowcolor{gray!8}
        \textbf{CUHK $\Rightarrow$ ICFG} & \textbf{Rank1} & \textbf{Rank5} & \textbf{Rank10}\\
        \hline
        Baseline & 30.79 & 49.10 & 57.47\\
        \rowcolor{lightkeycolor}
        Baseline+MCM & \textbf{31.13} & \textbf{49.52} & \textbf{58.07}\\
        \hline
        \textbf{ICFG $\Rightarrow$ CUHK} & Rank1 & Rank5 & Rank10\\
        \hline
        Baseline & 22.19 & 40.55 & 50.52\\
        \rowcolor{lightkeycolor}
        Baseline+MCM & \textbf{23.78} & \textbf{42.41} & \textbf{52.11}\\
        \hline
        \textbf{CUHK $\Rightarrow$ RSTP} & Rank1 & Rank5 & Rank10\\
        \hline
        Baseline & 37.40 & 63.40 & 74.35\\
        \rowcolor{lightkeycolor}
        Baseline+MCM & \textbf{39.25} & \textbf{63.95} & \textbf{74.55}\\
        \hline
        \textbf{RSTP $\Rightarrow$ CUHK} & Rank1 & Rank5 & Rank10\\
        \hline
        Baseline & 10.02 & 22.95 & 31.04\\
        \rowcolor{lightkeycolor}
        Baseline+MCM & \textbf{10.69} & \textbf{25.20} & \textbf{33.93}\\
        \bottomrule
    \end{tabular}}}
    \vspace{-15px}
\end{table}

To better understand the contributions of MCM in our framework, we conduct a domain generalization analysis on the three benchmarks to demonstrate our generalization performance. In Table~\ref{tab:e6}, CUHK $\Rightarrow$ ICFG and CUHK $\Rightarrow$ RSTP indicate using the model trained on the CUHK-PEDES dataset to infer their test sets and vice versa. Compared to the baseline, we can observe a performance improvement in the three metrics for ICFG-PEDES and RSTPReid. The fine-grained features mined by MCM are more resistant to overfitting. People-centric domain generalization provides related evidence that human-centered representation models should maintain discriminative behavior when data distributions shift across domains~\cite{liu2025domain}. The improvement in domain generalization shows the capability of our framework in learning generic cross-modal information, which is essential to solving the fine-grained modal heterogeneity.

\subsubsection{Effect of Margin Parameters}
\begin{figure}[!t]
  \centering
  \centerline{\includegraphics[width=0.70\textwidth]{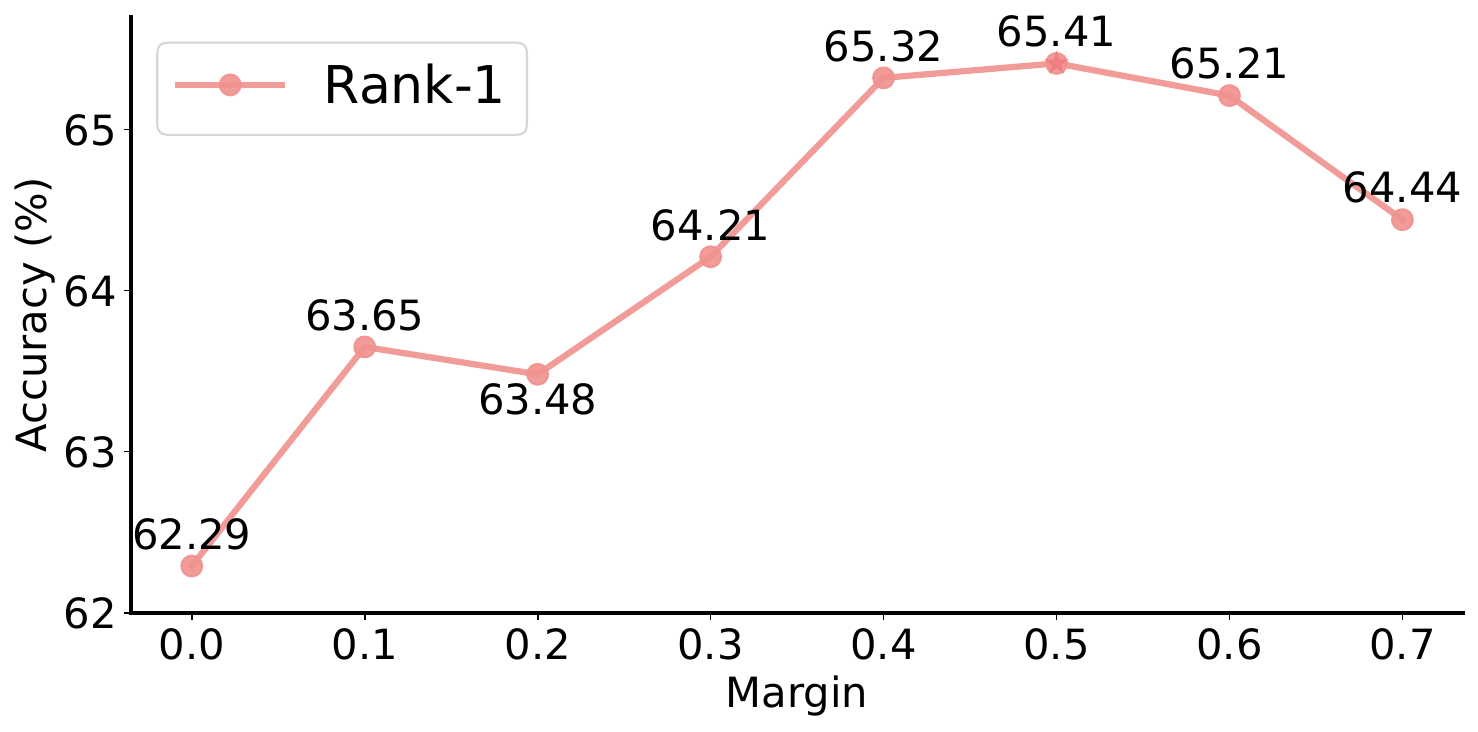}}
  \caption{Effect of the manual fixed margin parameter settings of our Sew Calibration loss in terms of Rank1 accuracy on the CUHK-PEDES.}
  \label{fig:margin}
\end{figure}

In Sec. ~\ref{sec:scl}, we introduce the adaptive margins to constrain cross-modal learning using Sew Calibration loss with constraints. We find that compared to the manually fixed parameters (SEW-f), our adaptive margins (SEW-a) produce better results, as shown in Fig.~\ref{fig:margin}, where we investigate the effect of manual margin in different settings on the CUHK-PEDES \cite{gnarnn} on Rank1 accuracy. The best performance of 65.41\% is achieved when $\mathcal{M}=0.5$, while the worst model performance of 62.29\% is observed when $\mathcal{M}=0$. We note that the performance first improves as the margin increases, which is due to our bi-directional margin compressing the same pedestrian target compactly. However, we find that the performance does not continue to improve as the margin increases, and we observe an inflection point around a margin of 0.5. When the margin is too large, the tight constraints make the model too hard to train.
Therefore, it is critical to utilize quality-guided adaptive margins according to the textual information. In this regard, we set the upper and lower margin bounds ($\mathcal{M}_{min}=0.4$ and $\mathcal{M}_{max}=0.6$). In this range, our Sew Calibration loss with adaptive constraints obtains the best performance, which a manually fixed margin cannot achieve, regardless of how the margin value is tuned.

\subsubsection{Effect of Mask Ratio}

Inspired by Masked Language Modeling \cite{lee2023lanobert}, we propose the MCM loss which leverages a masked captions prediction task to establish detailed and generic relationships between textual and visual parts. Our method requires setting the mask ratio in the MCM. To investigate the effect of the mask ratio on model performance and generalization learning ability, we conducted ablation experiments, and the results are shown in Table~\ref{tab:mr}. When the mask ratio is set to 0, which is equivalent to the baseline, the performance is the worst. Compared to the baseline, our method with a 0.1 mask ratio achieves the best results, with scores of 65.16\%, 81.95\%, and 87.93\% on Rank1, Rank5, and Rank10, respectively.
\begin{table}[!t]
    \centering
    \caption{Ablation results of mask ratio on the CUHK-PEDES.}
    \label{tab:mr}
    \setlength{\tabcolsep}{4.8mm}{
    \resizebox{.65\textwidth}{!}{
    \begin{tabular}{c|ccc}
        \toprule
        \rowcolor{gray!8}
        Baseline+MCM & \textbf{Rank1} & \textbf{Rank5} & \textbf{Rank10}\\
        \hline
        Mask = 0 & 62.29 & 81.25 & 87.22\\
        \hline
        \rowcolor{lightkeycolor}
        Mask = 0.1 & \textbf{65.16} & \textbf{81.95} & \textbf{87.93}\\
        \hline
        Mask = 0.3 & 64.97 & 81.38 & 87.71\\
        \hline
        Mask = 0.5 & 63.78 & 81.11 & 87.43\\
        \bottomrule
    \end{tabular}}}
    \vspace{-10px}
\end{table}
The experiments are conducted on the CUHK-PEDES dataset using the Baseline+MCM method. As shown in these results, the model achieves the best performance when the mask ratio is set to 0.1, and the performance gradually decreases as the ratio increases. We therefore keep 0.1 as a single shared hyperparameter on ICFG-PEDES and RSTPReID instead of re-tuning it per dataset, so the cross-dataset comparisons reflect the same MCM configuration. The main reason is that the information provided in the annotated caption text is already accurate and relatively concise. As the mask ratio expands, more key information may be obscured, causing the model to fail to learn enough semantic information to complete the construction of the image-text relationship.

\FloatBarrier
\subsection{Qualitative Analysis}
\subsubsection{Visualization of Attention Map}
\begin{figure}[!t]
  \centering
  \centerline{\includegraphics[width=0.88\textwidth]{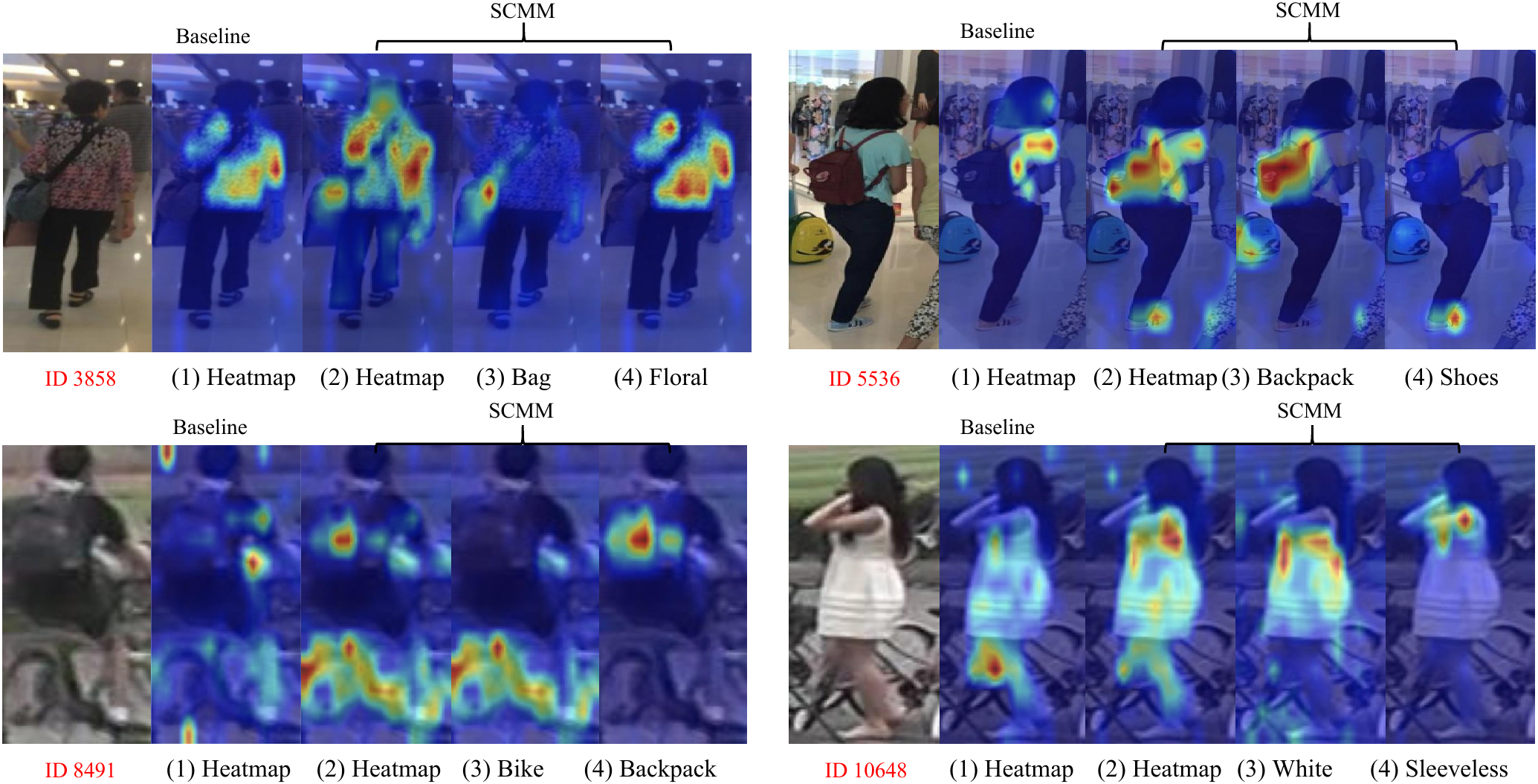}}
  \caption{Visualization of attention maps from baseline and SCMM on the CUHK-PEDES. We present the total caption-level results and the fine-grained word-level results, respectively. The caption-level heat maps average the last-layer attention over heads, while each word-level map uses the selected token as the query and averages its last-layer cross-attention over heads without averaging across different keywords. \textbf{(1)} Baseline caption-level results, \textbf{(2)} Our caption-level results, \textbf{(3) \& (4)} Our word-level results. Best viewed in color.}
  \label{fig:fig5}
\end{figure}
We visualize attention maps on the CUHK-PEDES test dataset to demonstrate model capability in learning image-text correspondence. As shown in Fig.~\ref{fig:fig5}, the visualization obtained by SCMM is more apparent and refined than the baseline. For the caption-level panels, the displayed response summarizes the final-layer attention pattern across heads. For the word-level panels, each heat map is generated from one selected keyword in the caption description, such as bag, shoes, colors, and clothing, and no additional averaging is performed across different tokens. For example, the visualizations of IDs 8491 and 10648 do not focus on the most useful detailed information under the baseline. The key messages in the two images are a man riding a bike with a backpack and a woman wearing a sleeveless white dress. SCMM captures these key details more reliably, showing that the MCM branch learns critical parts in the cross-modal correspondence.

\begin{figure}[!t]
  \centerline{\includegraphics[width=0.70\textwidth]{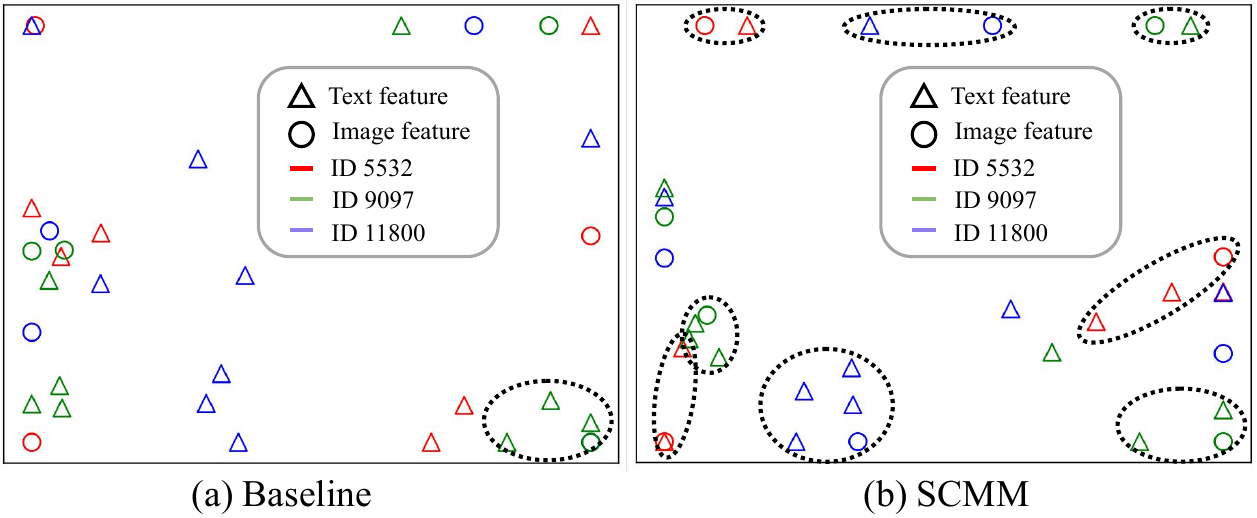}}
  \caption{Presentation of cross-modal representation distributions on CUHK-PEDES test dataset. Different colors correspond to the different target ID. Best viewed in color.}
  \label{fig:fig4}
\end{figure}

\subsubsection{Visualization of Image-text Feature Distribution}
SCMM is capable of learning better image-text representation distributions with fine-grained image-text interaction and adaptive constraints to reduce the cross-modal gap. To demonstrate this, we randomly selected some pedestrians and extracted their image-text global features, then mapped them to two dimensions for visualization in Fig.~\ref{fig:fig4}(a) and (b). For instance, we can take IDs 5532 and 11800 as examples. We can observe that the image-text features of the same ID distribution are compressed more compactly, and the boundaries between them are more apparent than the baseline results, which demonstrates the effectiveness of SCMM in mitigating the cross-modal gap.

\subsubsection{Visualization of Top-5 Retrieval Results}
\begin{figure}[!t]
  \centering
  \centerline{\includegraphics[width=0.7\textwidth]{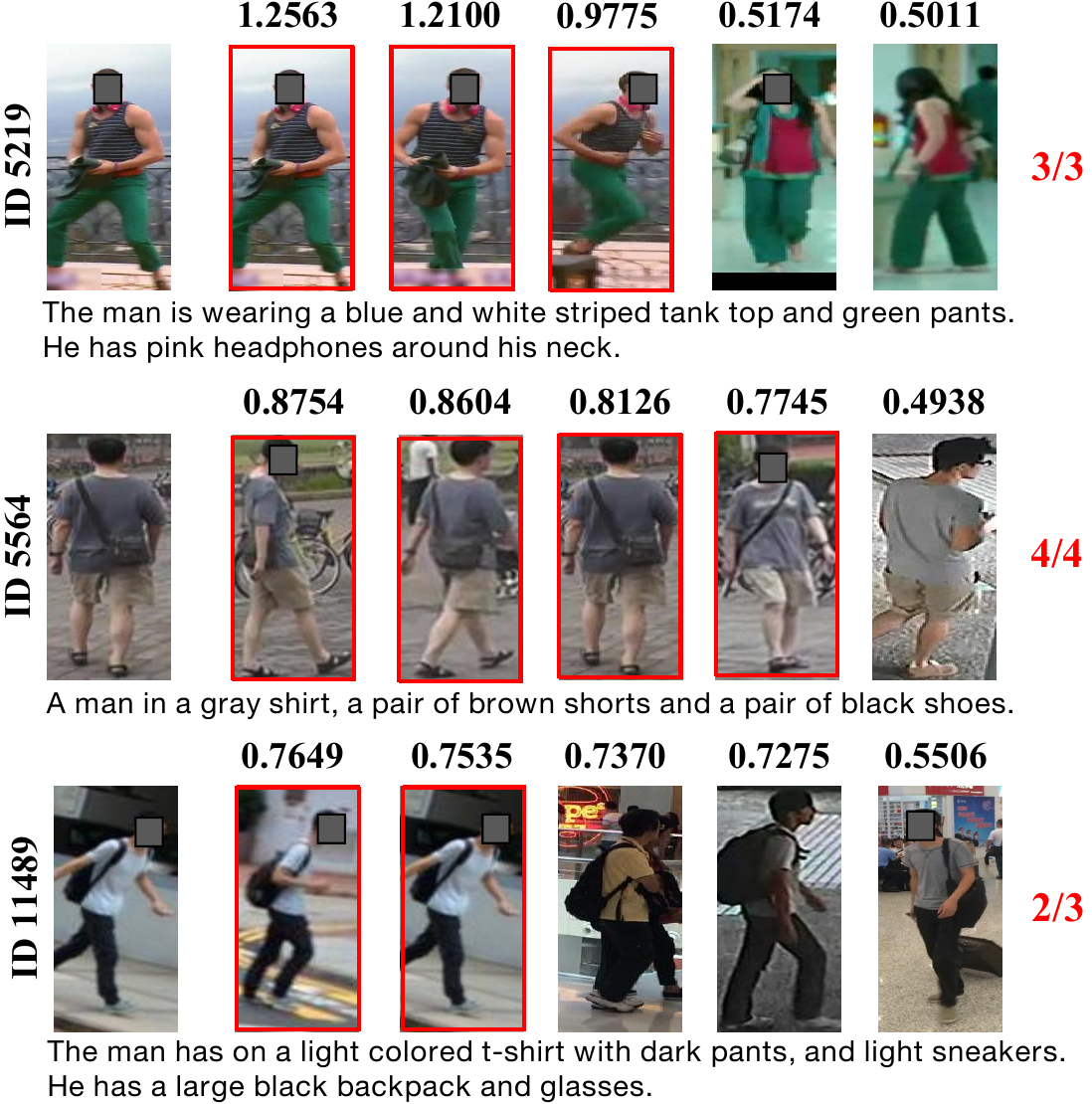}}
  \caption{Top-5 retrieval results on CUHK-PEDES. Matched target person images are marked with red rectangles. The subcaption below each image shows the text query used during retrieval. Best viewed in color.}
  \label{fig:visual}
\end{figure}

To comprehensively evaluate the retrieval performance of our method, we randomly selected several sets of search results and visualized them in Fig.~\ref{fig:visual}. For each target image, we show the top-5 most similar search results, with similarity scores after re-ranking displayed above each image. Correct retrievals are indicated by red rectangles, and the number in the last column shows how many pedestrian targets sharing an identical ID have been recalled.

Taking ID~5219 as an example, our method correctly retrieved all three target images, with similarity scores of 1.2563, 1.2100, and 0.9775, respectively. The two unmatched images received substantially lower similarity scores of 0.5174 and 0.5011. Although those images share partial visual similarities with the positive samples, they conflict with key details in the text description, demonstrating that SCMM has learned to differentiate subtle distinctions. The final row of Fig.~\ref{fig:visual} illustrates a particularly challenging case where the query person possesses highly similar attributes to multiple other pedestrians; nevertheless, SCMM correctly ranks two out of three positive images as the top-2 results.

\FloatBarrier
\section{Conclusion}
\label{sec:concl}

In this paper, we proposed SCMM, an innovative multimodal perception framework designed to calibrate cross-modal representations for text-based person search, advancing multimodal pattern recognition research. Focusing on the precise alignment of visual and textual semantics, SCMM uniquely integrates a sew calibration loss anchored by quality-guided adaptive margins, establishing global semantic alignment, and a masked caption modeling loss that facilitates complex fine-grained concept correspondence. Crucially, our dual-encoder architecture employs a training-only cross-modal decoder, thereby avoiding computational overhead during inference while maximizing cross-modal pattern discovery capabilities. Extensive experiments across three large-scale benchmark datasets consistently demonstrate that SCMM significantly advances the state-of-the-art framework boundaries, achieving 73.81\% Rank-1 retrieval accuracy on CUHK-PEDES, validating the effectiveness of our approach for multimodal perception and visual pattern recognition.

Despite these strengths, several limitations merit acknowledgment. First, the quality-guided adaptive margin currently uses text token length as a proxy for informational density, so it cannot explicitly distinguish concise but highly discriminative captions from long descriptions containing redundant phrases. Richer quality signals based on semantic complexity, attribute diversity, or encoder confidence may yield further gains. Second, SCMM requires fully annotated image--text pairs during training, which limits scalability to weakly supervised or cross-lingual retrieval scenarios where matched pairs are scarce. Third, the framework's discriminative capacity may diminish under severe occlusion or drastic appearance changes across non-overlapping camera views, an open challenge shared across TBPS methods.
The broader pattern recognition value of SCMM lies in its modular and efficient alignment design: the dual-encoder backbone remains suitable for large-scale retrieval, while sew calibration and MCM can be adapted to related cross-modal frameworks without adding inference-time cost. The domain-generalization results in Table~\ref{tab:e6} further indicate that MCM learns transferable cross-modal information across CUHK, ICFG, and RSTP. Future work will therefore focus on richer semantic quality estimators beyond token length, weaker supervision for scarce paired data, and extensions to video-based or open-vocabulary person retrieval.

\section*{CRediT authorship contribution statement}

\textbf{Jing Liu}: Conceptualization, Methodology, Formal analysis, Writing -- original draft, Writing -- review \& editing, Visualization. \textbf{Donglai Wei}: Methodology, Software, Validation, Formal analysis, Investigation, Writing -- review \& editing, Project administration. \textbf{Yang Liu}: Conceptualization, Methodology, Investigation, Writing -- review \& editing, Visualization. \textbf{Sipeng Zhang}: Writing -- review \& editing, Funding acquisition. \textbf{Tong Yang}: Writing -- review \& editing, Funding acquisition. \textbf{Wei Zhou}: Conceptualization, Writing -- review \& editing. \textbf{Weiping Ding}: Methodology, Writing -- review \& editing. \textbf{Victor C. M. Leung}: Supervision, Writing -- review \& editing, Project administration.

\section*{Data availability}

Data will be made available on request.

\section*{Declaration of competing interest}

The authors declare that they have no known competing financial interests or personal relationships that could have appeared to influence the work reported in this paper.

\section*{Declaration of Generative AI and AI-assisted technologies in the writing process}

During the preparation of this work, the authors used generative AI and AI-assisted technologies for writing enhancement and language polishing to improve the clarity and readability of the manuscript. After using these tools, the authors have reviewed and edited the content as needed and take full responsibility for the content of the publication.

\section*{Acknowledgements}

This work was supported in part by the Mitacs Elevate under Grant No. IT44479; the Guangdong Pearl River Talents Recruitment Program under Grant No. 2019ZT08X603; and the Guangdong Pearl Rivers Talent Plan under Grant No. 2019JC01X235. We sincerely thank all the editors and anonymous reviewers for their careful work and thoughtful suggestions that have helped improve this paper substantially.

\bibliographystyle{elsarticle-num}
\begin{spacing}{0.95}
  \bibliography{refs.bib}

@inproceedings{ren2022global,
  title     = {Global Optimal K-Medoids Clustering of One Million Samples},
  booktitle = {NeurIPS},
  author    = {Ren, Jiayang and Hua, Kaixun and Cao, Yankai},
  year      = 2022,
  volume    = {35},
  pages     = {982--994}
}

@misc{ren2023gpuaccelerated,
  title        = {A GPU-Accelerated Moving-Horizon Algorithm for Training Deep Classification Trees on Large Datasets},
  author       = {Ren, Jiayang and {Osuna-Enciso}, Valent{\'i}n and Okamoto, Morimasa and Mao, Qiangqiang and Ji, Chaojie and Cao, Liang and Hua, Kaixun and Cao, Yankai},
  year         = 2023,
  month        = nov,
  number       = {arXiv:2311.06952},
  eprint       = {2311.06952},
  primaryclass = {cs.LG},
  doi          = {10.48550/arXiv.2311.06952}
}

@article{ren2026global,
  title    = {A Global Optimization Algorithm for K-Center Clustering of One Billion Samples},
  author   = {Ren, Jiayang and You, Ningning and Hua, Kaixun and Ji, Chaojie and Cao, Yankai},
  year     = 2026,
  month    = jun,
  journal  = {Management Science},
  volume   = {72},
  number   = {6},
  pages    = {5106--5130},
  issn     = {0025-1909},
  doi      = {10.1287/mnsc.2023.00218}
}

@article{cao2025comprehensive,
  title    = {Comprehensive Analysis on Machine Learning Approaches for Interpretable and Stable Soft Sensors},
  author   = {Cao, Liang and Wang, Jingyi and Su, Jianping and Luo, Yi and Cao, Yankai and Braatz, Richard D. and Gopaluni, Bhushan},
  year     = 2025,
  journal = {IEEE Trans. Instrum. Meas.},
  volume   = {74},
  pages    = {1--17},
  issn     = {0018-9456, 1557-9662},
  doi      = {10.1109/TIM.2025.3556830}
}

@article{cao2025novel,
  title    = {A Novel Automated Soft Sensor Design Tool for Industrial Applications Based on Machine Learning},
  author   = {Cao, Liang and Su, Jianping and Conde, Emilio and Siang, Lim C. and Cao, Yankai and Gopaluni, Bhushan},
  year     = 2025,
  month    = jul,
  journal = {Control Eng. Pract.},
  volume   = {160},
  pages    = {106322},
  issn     = {09670661},
  doi      = {10.1016/j.conengprac.2025.106322}
}

@misc{liu2025anomaly,
  title        = {Anomaly Detection and Generation with Diffusion Models: A Survey},
  shorttitle   = {Anomaly Detection and Generation with Diffusion Models},
  author       = {Liu, Yang and Liu, Jing and Li, Chengfang and Xi, Rui and Li, Wenchao and Cao, Liang and Wang, Jin and Yang, Laurence T. and Yuan, Junsong and Zhou, Wei},
  year         = 2025,
  month        = jun,
  number       = {arXiv:2506.09368},
  eprint       = {2506.09368},
  primaryclass = {cs},
  doi          = {10.48550/arXiv.2506.09368}
}

@misc{wu2025survey,
  title        = {A Survey on Cloud-Edge-Terminal Collaborative Intelligence in AIoT Networks},
  author       = {Wu, Jiaqi and Liu, Jing and Liu, Yang and Wang, Lixu and Wang, Zehua and Chen, Wei and Tian, Zijian and Yu, Richard and Leung, Victor C. M.},
  year         = 2025,
  month        = aug,
  number       = {arXiv.2508.18803},
  eprint       = {2508.18803},
  primaryclass = {cs},
  doi          = {10.48550/arXiv.2508.18803}
}

@misc{liu2025multimodala,
  title      = {Multimodal Large Language Models in Medicine and Nursing: A Survey},
  shorttitle = {Multimodal Large Language Models in Medicine and Nursing},
  author     = {Liu, Jing and Gong, Linxiao and Guo, Juncen and Wu, Jingyi and Sun, Lianlong and Bi, Yulai and Patwari, Kartik and Chen, Boan and Zhang, Lichi and Zhou, Wei and Liu, Yang and Zhu, Xiaoguang and Chuah, Chen-Nee and Rajaratnam, Bala},
  year       = 2025,
  number     = {techrxiv.175623882.21520632/v1},
  doi        = {10.36227/techrxiv.175623882.21520632/v1}
}

@article{liu2025networking,
  title      = {Networking Systems for Video Anomaly Detection: A Tutorial and Survey},
  shorttitle = {Networking Systems for Video Anomaly Detection},
  author     = {Liu, Jing and Liu, Yang and Lin, Jieyu and Li, Jielin and Cao, Liang and Sun, Peng and Hu, Bo and Song, Liang and Boukerche, Azzedine and Leung, Victor C.M.},
  year       = 2025,
  month      = apr,
  journal    = {ACM Comput. Surv.},
  volume     = {57},
  number     = {10},
  pages      = {270:1--270:37},
  issn       = {0360-0300},
  doi        = {10.1145/3729222}
}

@inproceedings{liu2026collaborative,
  title     = {Collaborative Adaptive Curriculum for Progressive Knowledge Distillation},
  booktitle = {IEEE ICME},
  author    = {Liu, Jing and Ma, Zhenchao and Yu, Han and Ju, Bobo and Yang, Wenliang and Li, Chengfang and Hu, Bo and Song, Liang},
  year      = 2026,
  doi       = {10.48550/arXiv.2603.20296}
}

@inproceedings{liu2026diffusionguided,
  title     = {Diffusion-Guided Semantic Consistency for Multimodal Heterogeneity},
  booktitle = {IEEE ICME},
  author    = {Liu, Jing and Guo, Zhengliang and Wang, Yan and Zhu, Xiaoguang and Du, Yao and Wang, Zehua and Leung, Victor C. M.},
  year      = 2026,
  doi       = {10.48550/arXiv.2603.19337}
}

@inproceedings{liu2026semantic,
  title     = {Semantic Prototype Learning for Cross-Domain Activity Recognition},
  booktitle = {IEEE ICME},
  author    = {Liu, Jing and Yu, Han and Fang, Gaoyun and Wu, Jingyi and Gong, Linxiao and Zhu, Xiaoguang and Liu, Yang and Sun, Peng},
  year      = 2026
}

@misc{liu2025survey,
  title        = {A Survey on Diffusion Models for Anomaly Detection},
  author       = {Liu, Jing and Ma, Zhenchao and Wang, Zepu and Liu, Yang and Wang, Zehua and Sun, Peng and Song, Liang and Hu, Bo and Leung, Victor C. M.},
  year         = 2025,
  month        = jan,
  number       = {arXiv:2501.11430},
  eprint       = {2501.11430},
  primaryclass = {cs},
  doi          = {10.48550/arXiv.2501.11430}
}

@article{liu2025domain,
  title   = {Domain Generalization with Semi-Supervised Learning for People-Centric Activity Recognition},
  author  = {Liu, Jing and Zhu, Wei and Li, Di and Hu, Xing and Song, Liang},
  year    = 2025,
  month   = jan,
  journal = {Sci. China Inf. Sci.},
  volume  = {68},
  number  = {1},
  pages   = {112103},
  issn    = {1674-733X, 1869-1919},
  doi     = {10.1007/s11432-022-3860-y}
}

@article{liu2026edgecloud,
  title      = {Edge-Cloud Collaborative Computing on Distributed Intelligence and Model Optimization: A Survey},
  shorttitle = {Edge-Cloud Collaborative Computing on Distributed Intelligence and Model Optimization},
  author     = {Liu, Jing and Du, Yao and Yang, Kun and Wu, Jiaqi and Wang, Yan and Hu, Xiping and Wang, Zehua and Liu, Yang and Sun, Peng and Boukerche, Azzedine and Leung, Victor C. M.},
  year       = 2026,
  journal = {IEEE Commun. Surveys Tuts.},
  volume     = {28},
  pages      = {5049--5080},
  issn       = {1553-877X},
  doi        = {10.1109/COMST.2026.3669216}
}

@inproceedings{vitaa,
  title        = {Vitaa: Visual-textual attributes alignment in person search by natural language},
  author       = {Wang, Zhe and Fang, Zhiyuan and Wang, Jun and Yang, Yezhou},
  booktitle    = {ECCV},
  pages        = {402--420},
  year         = {2020},
  organization = {Springer}
}

@inproceedings{mgel,
  title     = {Text-based Person Search via Multi-Granularity Embedding Learning.},
  author    = {Wang, Chengji and Luo, Zhiming and Lin, Yaojin and Li, Shaozi},
  booktitle = {IJCAI},
  pages     = {1068--1074},
  year      = {2021}
}

@article{ssan,
  title   = {Semantically self-aligned network for text-to-image part-aware person re-identification},
  author  = {Ding, Zefeng and Ding, Changxing and Shao, Zhiyin and Tao, Dacheng},
  journal = {arXiv preprint arXiv:2107.12666},
  year    = {2021}
}

@article{nafs,
  title   = {Contextual non-local alignment over full-scale representation for text-based person search},
  author  = {Gao, Chenyang and Cai, Guanyu and Jiang, Xinyang and Zheng, Feng and Zhang, Jun and Gong, Yifei and Peng, Pai and Guo, Xiaowei and Sun, Xing},
  journal = {arXiv preprint arXiv:2101.03036},
  year    = {2021}
}

@article{tipcb,
  title     = {TIPCB: A simple but effective part-based convolutional baseline for text-based person search},
  author    = {Chen, Yuhao and Zhang, Guoqing and Lu, Yujiang and Wang, Zhenxing and Zheng, Yuhui},
  journal = {Neurocomputing},
  volume    = {494},
  pages     = {171--181},
  year      = {2022},
  publisher = {Elsevier}
}

@inproceedings{safaNet,
  title        = {Learning Semantic-Aligned Feature Representation for Text-Based Person Search},
  author       = {Li, Shiping and Cao, Min and Zhang, Min},
  booktitle    = {ICASSP},
  pages        = {2724--2728},
  year         = {2022},
  organization = {IEEE}
}

@inproceedings{ivt,
  title        = {See finer, see more: Implicit modality alignment for text-based person retrieval},
  author       = {Shu, Xiujun and Wen, Wei and Wu, Haoqian and Chen, Keyu and Song, Yiran and Qiao, Ruizhi and Ren, Bo and Wang, Xiao},
  booktitle    = {ECCV Wkshps},
  pages        = {624--641},
  year         = {2023},
  organization = {Springer}
}

@inproceedings{cmpmcmpc,
  title     = {Deep cross-modal projection learning for image-text matching},
  author    = {Zhang, Ying and Lu, Huchuan},
  booktitle = {ECCV},
  pages     = {686--701},
  year      = {2018}
}

@article{dualpath,
  title     = {Dual-path convolutional image-text embeddings with instance loss},
  author    = {Zheng, Zhedong and Zheng, Liang and Garrett, Michael and Yang, Yi and Xu, Mingliang and Shen, Yi-Dong},
  journal   = {ACM Trans. Multimedia Comput. Commun. Appl.},
  volume    = {16},
  number    = {2},
  pages     = {1--23},
  year      = {2020},
  publisher = {ACM New York, NY, USA}
}

@article{alignbefore,
  title   = {Align before fuse: Vision and language representation learning with momentum distillation},
  author  = {Li, Junnan and Selvaraju, Ramprasaath and Gotmare, Akhilesh and Joty, Shafiq and Xiong, Caiming and Hoi, Steven Chu Hong},
  journal = {NeurIPS},
  volume  = {34},
  pages   = {9694--9705},
  year    = {2021}
}

@inproceedings{caibc,
  title     = {CAIBC: Capturing All-round Information Beyond Color for Text-based Person Retrieval},
  author    = {Wang, Zijie and Zhu, Aichun and Xue, Jingyi and Wan, Xili and Liu, Chao and Wang, Tian and Li, Yifeng},
  booktitle = {ACM MM},
  pages     = {5314--5322},
  year      = {2022}
}

@inproceedings{dssl,
  title     = {DSSL: Deep Surroundings-person Separation Learning for Text-based Person Retrieval},
  author    = {Zhu, Aichun and Wang, Zijie and Li, Yifeng and Wan, Xili and Jin, Jing and Wang, Tian and Hu, Fangqiang and Hua, Gang},
  booktitle = {ACM MM},
  pages     = {209--217},
  year      = {2021}
}

@article{isa,
  title   = {Image-Specific Information Suppression and Implicit Local Alignment for Text-Based Person Search},
  author  = {Yan, Shuanglin and Tang, Hao and Zhang, Liyan and Tang, Jinhui},
  year    = {2024},
  journal = {IEEE Trans. Neural Netw. Learn. Syst.},
  volume  = {35},
  number  = {12},
  pages   = {17973--17986},
  issn    = {2162-2388}
}

@inproceedings{gnarnn,
  title     = {Person search with natural language description},
  author    = {Li, Shuang and Xiao, Tong and Li, Hongsheng and Zhou, Bolei and Yue, Dayu and Wang, Xiaogang},
  booktitle = {CVPR},
  pages     = {1970--1979},
  year      = {2017}
}

@article{axm,
  title   = {AXM-Net: cross-modal context sharing attention network for person Re-ID},
  author  = {Farooq, Ammarah and Awais, Muhammad and Kittler, Josef and Khalid, Syed Safwan},
  journal = {arXiv preprint arXiv:2101.08238},
  year    = {2021}
}

@article{verbalNet,
  title     = {Verbal-Person Nets: Pose-Guided Multi-Granularity Language-to-Person Generation},
  author    = {Liu, Deyin and Wu, Lin and Zheng, Feng and Liu, Lingqiao and Wang, Meng},
  journal   = {IEEE Trans. Neural Netw. Learn. Syst.},
  year      = {2022},
  publisher = {IEEE}
}

@inproceedings{circle,
  title     = {Circle loss: A unified perspective of pair similarity optimization},
  author    = {Sun, Yifan and Cheng, Changmao and Zhang, Yuhan and Zhang, Chi and Zheng, Liang and Wang, Zhongdao and Wei, Yichen},
  booktitle = {CVPR},
  pages     = {6398--6407},
  year      = {2020}
}

@article{vit,
  title   = {An image is worth 16x16 words: Transformers for image recognition at scale},
  author  = {Dosovitskiy, Alexey and Beyer, Lucas and Kolesnikov, Alexander and Weissenborn, Dirk and Zhai, Xiaohua and Unterthiner, Thomas and Dehghani, Mostafa and Minderer, Matthias and Heigold, Georg and Gelly, Sylvain and others},
  journal = {arXiv preprint arXiv:2010.11929},
  year    = {2020}
}

@inproceedings{imagenet,
  title        = {Imagenet: A large-scale hierarchical image database},
  author       = {Deng, Jia and Dong, Wei and Socher, Richard and Li, Li-Jia and Li, Kai and Fei-Fei, Li},
  booktitle    = {CVPR},
  pages        = {248--255},
  year         = {2009},
  organization = {IEEE}
}

@inproceedings{bai2023textbased,
  title     = {Text-Based Person Search without Parallel Image-Text Data},
  booktitle = {ACM MM},
  author    = {Bai, Yang and Wang, Jingyao and Cao, Min and Chen, Chen and Cao, Ziqiang and Nie, Liqiang and Zhang, Min},
  year      = {2023},
  pages     = {757--767},
  isbn      = {9798400701085}
}

@inproceedings{jiang2023crossmodal,
  title     = {Cross-Modal Implicit Relation Reasoning and Aligning for Text-to-Image Person Retrieval},
  booktitle = {CVPR},
  author    = {Jiang, Ding and Ye, Mang},
  year      = {2023},
  pages     = {2787--2797},
  isbn      = {9798350301298}
}

@article{cfine,
  title   = {CLIP-Driven Fine-Grained Text-Image Person Re-Identification},
  author  = {Yan, Shuanglin and Dong, Neng and Zhang, Liyan and Tang, Jinhui},
  year    = {2023},
  journal = {IEEE Trans. Image Process.},
  volume  = {32},
  pages   = {6032--6046},
  issn    = {1941-0042}
}

@article{sum,
  title     = {SUM: Serialized Updating and Matching for text-based person retrieval},
  author    = {Wang, Zijie and Zhu, Aichun and Xue, Jingyi and Jiang, Daihong and Liu, Chao and Li, Yifeng and Hu, Fangqiang},
  journal   = {Knowl. Based Syst.},
  volume    = {248},
  pages     = {108891},
  year      = {2022},
  publisher = {Elsevier}
}

@article{acsa,
  title     = {Asymmetric Cross-Scale Alignment for Text-Based Person Search},
  author    = {Ji, Zhong and Hu, Junhua and Liu, Deyin and Wu, Lin Yuanbo and Zhao, Ye},
  journal   = {IEEE Trans. Multimedia},
  year      = {2022},
  publisher = {IEEE}
}

@article{limit,
  title   = {Text-based person search with limited data},
  author  = {Han, Xiao and He, Sen and Zhang, Li and Xiang, Tao},
  journal = {arXiv preprint arXiv:2110.10807},
  year    = {2021}
}

@inproceedings{clip,
  title        = {Learning transferable visual models from natural language supervision},
  author       = {Radford, Alec and Kim, Jong Wook and Hallacy, Chris and Ramesh, Aditya and Goh, Gabriel and Agarwal, Sandhini and Sastry, Girish and Askell, Amanda and Mishkin, Pamela and Clark, Jack and others},
  booktitle    = {ICML},
  pages        = {8748--8763},
  year         = {2021},
  organization = {PMLR}
}

@inproceedings{facenet,
  title     = {Facenet: A unified embedding for face recognition and clustering},
  author    = {Schroff, Florian and Kalenichenko, Dmitry and Philbin, James},
  booktitle = {CVPR},
  pages     = {815--823},
  year      = {2015}
}

@inproceedings{zhai2022lit,
  title     = {Lit: Zero-shot transfer with locked-image text tuning},
  author    = {Zhai, Xiaohua and Wang, Xiao and Mustafa, Basil and Steiner, Andreas and Keysers, Daniel and Kolesnikov, Alexander and Beyer, Lucas},
  booktitle = {CVPR},
  pages     = {18123--18133},
  year      = {2022}
}

@article{devlin2018bert,
  title   = {Bert: Pre-training of deep bidirectional transformers for language understanding},
  author  = {Devlin, Jacob and Chang, Ming-Wei and Lee, Kenton and Toutanova, Kristina},
  journal = {arXiv preprint arXiv:1810.04805},
  year    = {2018}
}

@article{attention,
  title   = {Attention is all you need},
  author  = {Vaswani, Ashish and Shazeer, Noam and Parmar, Niki and Uszkoreit, Jakob and Jones, Llion and Gomez, Aidan N and Kaiser, {\L}ukasz and Polosukhin, Illia},
  journal = {NeurIPS},
  volume  = {30},
  year    = {2017}
}

@article{gong2024crossmodal,
  title   = {Cross-Modal Semantic Aligning and Neighbor-Aware Completing for Robust Text--Image Person Retrieval},
  author  = {Gong, Tiantian and Wang, Junsheng and Zhang, Liyan},
  year    = {2024},
  journal = {Inf. Fusion},
  volume  = {112},
  pages   = {102544},
  issn    = {15662535}
}

@article{ding2025dynamic,
  title   = {Dynamic Evidence Fusion Neural Networks with Uncertainty Theory and Its Application in Brain Network Analysis},
  author  = {Ding, Weiping and Hou, Tao and Huang, Jiashuang and Ju, Hengrong and Jiang, Shu},
  year    = {2025},
  journal = {Inf. Sci.},
  volume  = {691},
  pages   = {121622},
  issn    = {00200255}
}

@article{qian2025survey,
  title      = {A Survey on Multi-View Fusion for Predicting Links in Biomedical Bipartite Networks: Methods and Applications},
  shorttitle = {A Survey on Multi-View Fusion for Predicting Links in Biomedical Bipartite Networks},
  author     = {Qian, Yuqing and Wang, Yizheng and Liu, Junkai and Zou, Quan and Ding, Yijie and Guo, Xiaoyi and Ding, Weiping},
  year       = {2025},
  journal    = {Inf. Fusion},
  volume     = {117},
  pages      = {102894},
  issn       = {15662535}
}

@article{lee2023lanobert,
  title      = {LAnoBERT: System Log Anomaly Detection Based on BERT Masked Language Model},
  shorttitle = {LAnoBERT},
  author     = {Lee, Yukyung and Kim, Jina and Kang, Pilsung},
  year       = {2023},
  journal    = {Appl. Soft Comput.},
  volume     = {146},
  pages      = {110689},
  issn       = {1568-4946}
}

@article{niu2024comprehensive,
  title   = {Comprehensive Attribute Prediction Learning for Person Search by Language},
  author  = {Niu, Kai and Huang, Linjiang and Long, Yuzhou and Huang, Yan and Wang, Liang and Zhang, Yanning},
  year    = {2024},
  journal = {IEEE Trans. Image Process.},
  volume  = {33},
  pages   = {1990--2003},
  issn    = {1941-0042},
  doi     = {10.1109/TIP.2024.3372832}
}

@article{zhang2025localenhanced,
  title   = {Local-Enhanced Representation for Text-Based Person Search},
  author  = {Zhang, Guoqing and Chen, Yuhao and Zheng, Yuhui and Martin, Gaven and Wang, Ruili},
  year    = {2025},
  journal = {Pattern Recognit.},
  volume  = {161},
  pages   = {111247},
  issn    = {0031-3203},
  doi     = {10.1016/j.patcog.2024.111247}
}

@article{xie2025multiscale,
  title   = {Multi-Scale Feature Refinement via Perspective Scaling and Adaptive Regularization for Text-Based Person Search},
  author  = {Xie, Sheng and Zhang, Canlong and Ma, Runcong and Li, Zhixin and Wang, Zhiwen and Wei, Chunrong},
  year    = {2025},
  journal = {Eng. Appl. Artif. Intell.},
  volume  = {159},
  pages   = {111496},
  issn    = {0952-1976},
  doi     = {10.1016/j.engappai.2025.111496}
}

@article{yu2025climbreid,
  title      = {CLIMB-ReID: A Hybrid CLIP-Mamba Framework for Person Re-Identification},
  shorttitle = {CLIMB-ReID},
  author     = {Yu, Chenyang and Liu, Xuehu and Zhu, Jiawen and Wang, Yuhao and Zhang, Pingping and Lu, Huchuan},
  year       = {2025},
  journal    = {AAAI},
  volume     = {39},
  number     = {9},
  pages      = {9589--9597},
  issn       = {2374-3468},
  doi        = {10.1609/aaai.v39i9.33039}
}

@article{yu2024discovering,
  title   = {Discovering attention-guided cross-modality correlation for visible-infrared person re-identification},
  author  = {Yu, Hao and Cheng, Xu and Cheng, Kevin Ho Man and Peng, Wei and Yu, Zitong and Zhao, Guoying},
  year    = {2024},
  journal = {Pattern Recognit.},
  volume  = {155},
  pages   = {110643},
  doi     = {10.1016/j.patcog.2024.110643}
}

@article{feng2025homogeneous,
  title   = {Homogeneous and heterogeneous relational graph for visible-infrared person re-identification},
  author  = {Feng, Yujian and Chen, Feng and Yu, Jian and Ji, Yimu and Wu, Fei and Liu, Shangdong and Jing, Xiao-Yuan},
  year    = {2025},
  journal = {Pattern Recognit.},
  volume  = {158},
  pages   = {110981},
  doi     = {10.1016/j.patcog.2024.110981}
}
\end{spacing}

\end{document}